\documentclass{bmvc2k}

\title{PT43D: A Probabilistic Transformer for \\ Generating 3D Shapes from \\ Single Highly-Ambiguous RGB Images}

\addauthor{Yiheng Xiong}{yiheng.xiong@tum.de}{1}
\addauthor{Angela Dai}{angela.dai@tum.de}{1}

\addinstitution{
 3D AI Lab\\
 Technical University of Munich\\
 Munich, Germany
}

\runninghead{Xiong and Dai}{PT43D}

\def\etal{\emph{et al}\bmvaOneDot}

\begin{document}

\maketitle

\begin{figure}[ht]
  \centering
  \includegraphics[width=1.0\textwidth]{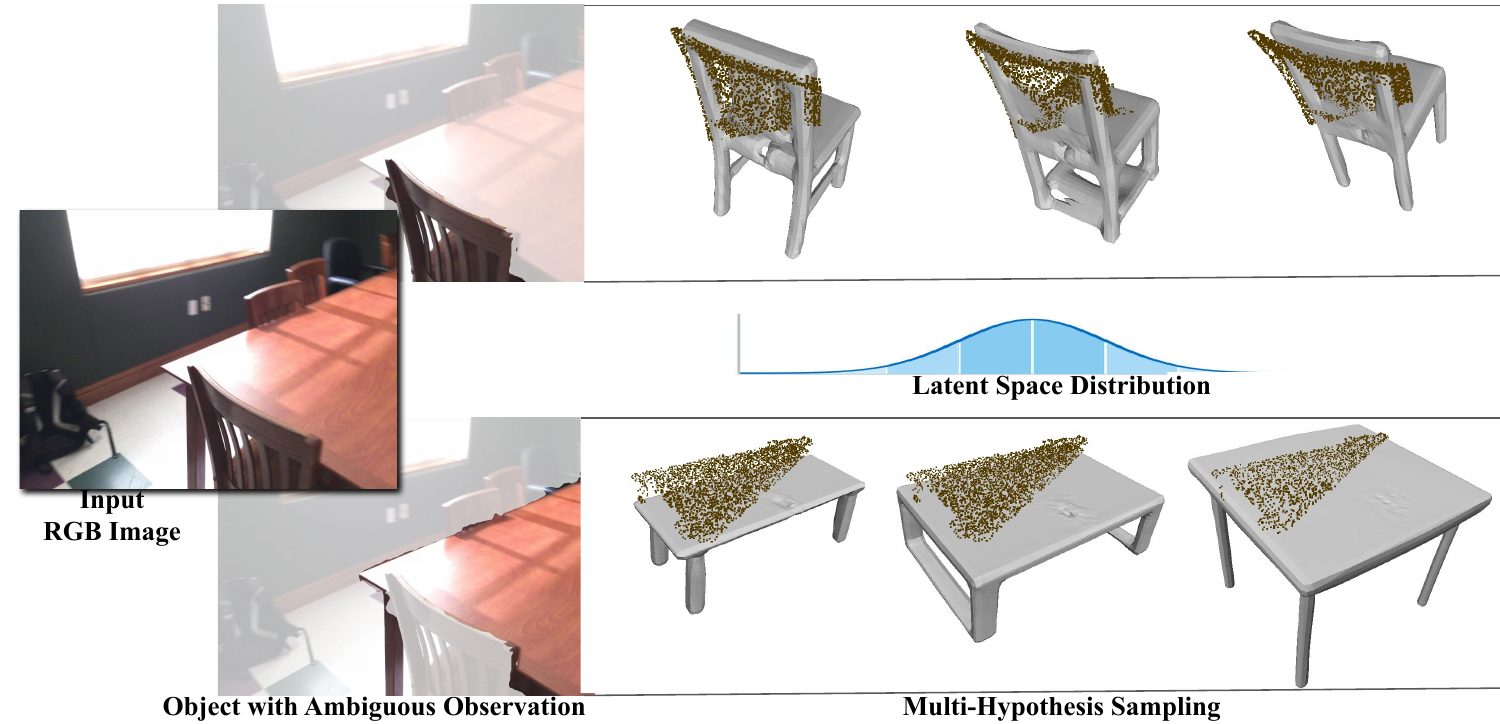}
  \caption{Our approach seeks to model the distribution of 3D shapes in latent space conditioned on a single highly ambiguous image, enabling the sampling of multiple diverse hypotheses during inference. Colored points are for visualization purposes only, indicating the overlap between ground-truth visible parts and plausible shapes.
  }
  \label{fig:teaser}
\end{figure}

\begin{abstract}
Generating 3D shapes from single RGB images is essential in various applications such as robotics. Current approaches typically target images containing clear and complete visual descriptions of the object, without considering common realistic cases where observations of objects that are largely occluded or truncated. We thus propose a transformer-based autoregressive model to generate the probabilistic distribution of 3D shapes conditioned on an RGB image containing potentially highly ambiguous observations of the object. To handle realistic scenarios such as occlusion or field-of-view truncation, we create simulated image-to-shape training pairs that enable improved fine-tuning for real-world scenarios. We then adopt cross-attention to effectively identify the most relevant region of interest from the input image for shape generation. This enables inference of sampled shapes with reasonable diversity and strong alignment with the input image. We train and test our model on our synthetic data then fine-tune and test it on real-world data. Experiments demonstrate that our model outperforms state of the art in both scenarios\footnote {Our code is open sourced at \href{https://github.com/xiongyiheng/PT43D}{https://github.com/xiongyiheng/PT43D}.}.

\end{abstract}
\section{Introduction}
\label{sec:introduction}
Generating 3D shapes from single RGB images presents a fundamental challenge in computer vision, with practical applications such as single-view 3D reconstruction, where the objective is to reconstruct the 3D structure and geometry of the objects in an observed image. To tackle this challenge, researchers have heavily relied on data-driven deep learning techniques. Earlier approaches~\cite{choy20163d,mescheder2019occupancy,richter2018matryoshka,xie2019pix2vox,xie2020pix2vox++,xu2019disn} focus on generating a deterministic 3D output based on an RGB image, achieving impressive results, but without considering inherent ambiguities that often arise from monocular perception. For example, an image of a chair from an upper view does not reveal the shape of its legs (see Fig.\ref{fig:teaser} top part). In contrast, recent advancements in generative AI~\cite{goodfellow2020generative,ho2020denoising,rombach2022high} have led to approaches~\cite{mittal2022autosdf,cheng2023sdfusion,vahdat2022lion,zheng2022sdf} that aim to learn the entire distribution of 3D shapes conditioned on the input image, enabling the sampling of multiple 3D hypotheses at test time. Nevertheless, these approaches target scenarios where images contain a fairly clear, complete visual description of the object, while many real-world scenarios contain observations of objects that are highly occluded or truncated. For instance, a home robot operating in cluttered environments may encounter situations where parts of objects are hidden from view due to limited field-of-view of the camera or obstacles in the scene. In such cases, relying solely on methods designed for complete and unobstructed observations may lead to inaccurate understanding of the environment, limiting the robot's ability to make informed decisions and navigate effectively.

To this end, we propose a transformer-based~\cite{vaswani2017attention} autoregressive model to generate the probabilistic distribution of 3D shapes conditioned on a single RGB image containing potentially highly ambiguous observations of the object. To handle realistic scenarios such as occlusion or field-of-view truncation, we create simulated image-to-shape training pairs from ShapeNet~\cite{chang2015shapenet} that enable improved fine-tuning for real-world scenarios. During training, each image can be mapped to potentially multiple ground-truth shapes due to inherent ambiguity in the input image observation. We then adopt cross-attention to effectively identify the most relevant region of interest from the input image for shape generation. This enables inference of sampled shapes that meet two fundamental criteria: reasonable diversity and strong alignment with the input image. We train and test our model on our synthetic training pairs then fine-tune and test it on real-world data from ScanNet~\cite{dai2017scannet,avetisyan2019scan2cad}. Experiments demonstrate that our model outperforms state of the art in both scenarios.

We summarize our contributions as follows:
\begin{itemize}
  \item We propose the first probabilistic approach for 3D shape generation from a highly ambiguous RGB image, leveraging cross-attention to effectively identify the most relevant region of interest from the input image, such that diverse output hypotheses all align well with the input image.
  \item We introduce a synthetic data augmentation approach to enable effective multi-hypothesis generation by setting one image maps to potentially multiple ground-truth shapes during training. Our approach with synthetic data enables pre-training that notably improves shape reconstruction results when fine-tuned on real-world data.
\end{itemize}
\section{Related Work}
\label{sec:realated_work}
\noindent \textbf{Single-View Object Reconstruction.} Inferring 3D shapes from monocular views is an inherently ill-posed task. Early approaches generate single 3D outputs from single RGB images in a deterministic manner, utilizing various representations such as voxels~\cite{choy20163d,xie2019pix2vox,xie2020pix2vox++,girdhar2016learning,wu2017marrnet,tulsiani2017multi,richter2018matryoshka}, point clouds~\cite{fan2017point,mandikal20183d,wu2020pq}, meshes~\cite{wang2018pixel2mesh,wang20193dn}, and CAD retrieval~\cite{kuo2020mask2cad,kuo2021patch2cad,gao2023diffcad}. 
More recent approaches also focus on implicit representations such as SDFs~\cite{jiang2020sdfdiff,xu2019disn}, UDFs~\cite{chibane2020neural}, and closest surface points~\cite{venkatesh2021deep}, achieving impressive reconstruction fidelity. However, these approaches do not address the inherent ambiguity in a single RGB image, which often results in the co-existence of multiple plausible reconstructions that explain the input image. Our method tackles this ambiguity by treating image-based predictions as conditional distributions, allowing for generation of multiple plausible sampled reconstructions.

\noindent \textbf{Probabilistic 3D Shape Generation.} More recently, researchers have explored the use of generative neural networks~\cite{ho2020denoising,rombach2022high,goodfellow2020generative,van2016pixel,van2016wavenet} to model the probabilistic distribution of 3D shapes, enabling the sampling of multiple hypotheses from the learned distribution. For instance, ~\cite{yan2022shapeformer,ibing2021octree,siddiqui2024meshgpt,nash2020polygen,zhang20223dilg} utilize transformer-based architecture to generate shapes, achieving impressive results. With recent advancements in denoising diffusion models, ~\cite{mo2024dit,hui2022neural,gupta20233dgen,zhang20233dshape2vecset} generate high-quality shapes by iteratively refining noise-corrupted data through a reverse diffusion process guided by learned probability distributions. For image-conditioned shape generation, AutoSDF~\cite{mittal2022autosdf} models shape distribution via an autoregressive process and combines it with a separately-trained image-conditioning model to generate shapes based on single RGB images. SDFusion~\cite{cheng2023sdfusion} learns to sample from a target shape distribution by reversing a progressive noise diffusion process, generating single-view reconstructions via classifier-free image guidance~\cite{ho2022classifier}. These approaches typically focus on images with fairly clear and complete visual descriptions of the object, but unfortunately do not address common real-world scenarios where objects are captured under strong occlusion or truncation. Our method, instead, takes advantage of generative neural networks, analyzing common realistic scenarios where images are highly occluded or truncated.
\section{Methodology}
\label{sec:methodology}
Our transformer-based approach seeks to model the distribution that represents potential 3D shapes conditioned on the input image. Given a shape $X$ represented via a T-SDF, we first follow the P-VQ-VAE~\cite{mittal2022autosdf} to compress it to a low-dimensional and discrete-grid representation $E$. Each grid holds a 3D latent feature index $q_i$ corresponding to the shape feature $e_i$ from a codebook $Z$ that is used for indexing feature embeddings. Then we learn the conditional distribution of shapes based on the input image $c$ over this compressed latent representation. 
\subsection{Shape Compression and Discretization}
For each shape $X$, a shape encoder $E_{\phi}(\cdot)$ encodes its Truncated Signed Distance Field (T-SDF) into a low-dimensional 3D grid feature $E \in g^3 \times D$, where $g$ represents the resolution of the 3D grid, and $D$ denotes the dimension of the 3D grid features. We establish a discrete latent space containing $K$ embedding vectors, each with a dimension of $D$. For the shape feature $e_i$ at grid position $i$, a vector quantization operation $VQ(\cdot)$ is employed to find its nearest neighbor $e'_i$ from the $K$ embedding vectors in the discrete latent space. The feature input into the shape decoder $D_{\phi}(\cdot)$ is the 3D discrete latent feature grid $E' \in g^3 \times D$, comprising discrete features $e'_i$ at grid locations i. The formula is:
\begin{equation}
    E = E_{\phi}(X), E' = VQ(E), X' = D_{\phi}(E').
\end{equation}
where $X'$ is the reconstructed shape, and $\phi$ is the set of learnable parameters of the decoder. As a result, a shape $X$ can be represented as a 3D latent feature index grid $Q\in g^3$, where $q_i$ at each location corresponds to a discrete shape feature $e'_i$. In particular, we adopt Patch-wised encoding VQ-VAE (P-VQ-VAE)~\cite{mittal2022autosdf} to learn this process since it allows for a better representation for local details. Given the 3D T-SDF $X$ of the shape, we first split $X$ into $N$ patches of $X_p$. Then $E_{\phi}$ encodes each patch independently and $D_{\phi}$ decodes all patches jointly. Empirically, we set $K = 512$, $D = 256$, $g = 8$ and $N = 512$. Training the network uses a combination of three losses proposed by van den Oord \etal ~\cite{oord2018neural}: reconstruction loss, the vector quantization objective and the commitment loss. 
\subsection{Method Overview} 
\begin{figure}[ht]
  \centering
  \includegraphics[width=0.95\textwidth]{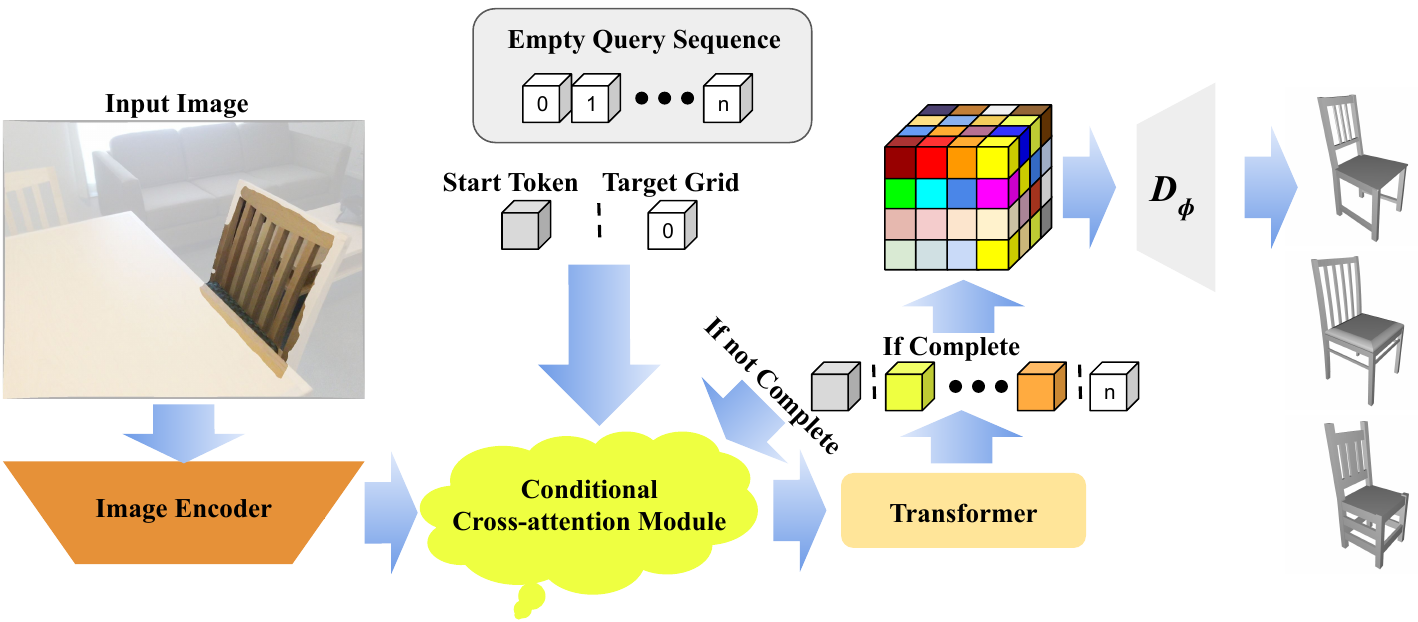}
  \vspace{-0.4cm}
  \caption{\textbf{Method Overview}. Our model adopts a low-dimensional and discretized latent space for shape generation. Given an input RGB image, it is first processed by an image encoder. Then the extracted image encodings are fed into the conditional module where they attend with the input sequence consisting of a start token, potentially filled grids containing latent features, and a target grid containing its position embedding via cross-attention such that the output sequence contains essential image features. Then the transformer processes this sequence and predicts the latent feature for the target grid. Conditional cross-attention and transformer are adopted autoregressively until a complete sequence is output where each cell contains the conditional distribution of latent features. During inference, we can gather the complete sequence into grids then sample and decode to multiple plausible hypotheses.
  } \label{fig:method_overview}
\end{figure}

Given an input image $c$, an empty query sequence $s_q$ in the length of $n + 1$ and a start token $<sos>$, our model generates the shape distribution over the latent space $E'$ by filling the query sequence autoregressively. The complete sequence $s_c$ comprises of discrete grids, where each location $i$ contains the distribution $p_i(e')$ of 3D latent features. During inference, we can sample from $p_i(e')$ to acquire shape feature index $q_i$ for each location and further get the corresponding shape feature $e'_i$ from the codebook $Z$. Finally, a pretrained decoder $D_\phi$ utilizes these shape features to produce the reconstruction output. The overall framework, as displayed in Fig~\ref{fig:method_overview}, comprises three key components: the image encoder, the conditional cross-attention module and the autoregressive modeling with transformer.

\noindent \textbf{Image Encoder.} The role of the image encoder is important in the context of image-based 3D shape generation. An effective image encoder possesses the capability to extract essential features from images, playing a crucial role in facilitating the subsequent 3D shape generation process. We adopt a pretrained "ViT-B/32" CLIP~\cite{radford2021learning} to extract the input image features, chosen for its robust capacity to capture rich image representations. 

\begin{figure}[ht]
    \centering
    \includegraphics[width=0.9\linewidth]{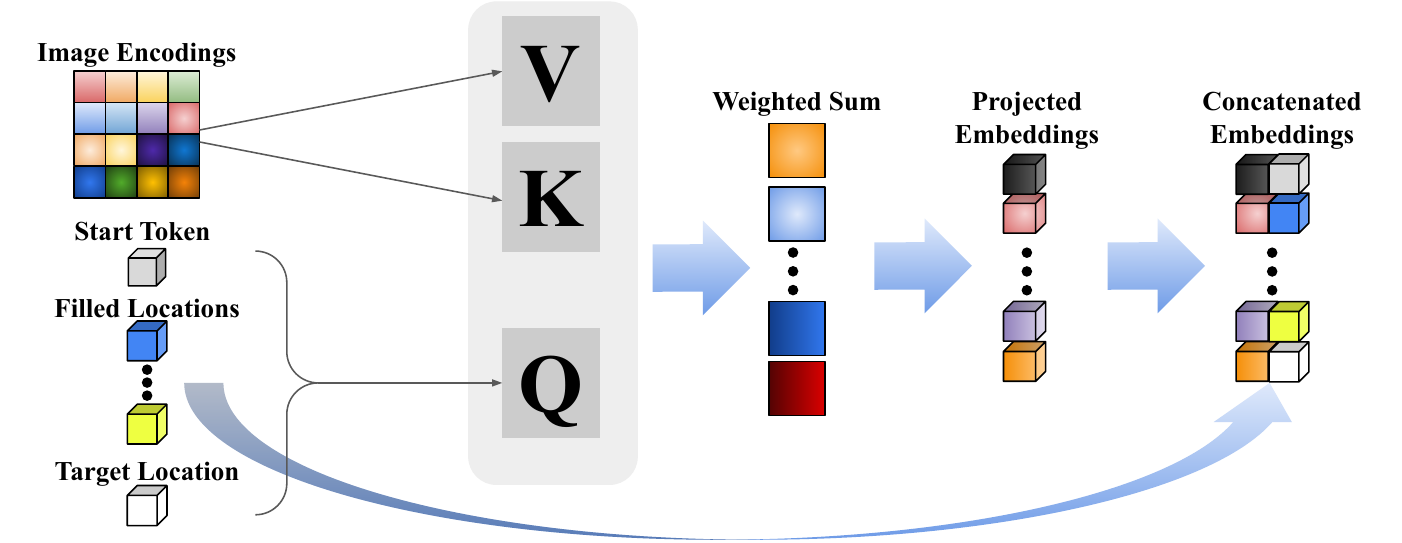}
  \vspace{-0.4cm}
    \caption{\textbf{Identifying Most Relevant Regions in the Image via Conditional Cross-Attention.}
    We perform cross-attention between image encodings and the input sequence to effectively identify the most relevant region of interest from the input image. The input sequence consisting of a start token, potentially filled locations with their own predicted latent features, and a target location with its position embedding, is multiplied with a weight matrix $Q$ and the image encodings are multiplied with weight matrices $V$ and $K$ independently. The resulting sequence contains a weighted sum of image features. We then project it to the original dimension and concatenate with the original sequence embeddings.
    }
    \label{fig:conditional_module}
\end{figure}

\noindent \textbf{Conditional Cross-Attention Module.} The conditional module establishes a connection between the image encodings and the grid sequence $s_t$ at current autoregressive step $t$, ensuring that the output grid sequence incorporates the essential image features for subsequent operations. Fig.~\ref{fig:conditional_module} depicts the architecture of it. $s_t$ comprises of the start token $<sos>$, potentially filled grids containing latent features, and the target grid with its position embedding. We perform cross-attention between $s_t$ and the image encodings, where each grid in $s_t$ serves as a query. The output sequence after cross-attention is essentially a weighted sum of image features. Then we project it to the original dimension and concatenate it with original embeddings. 

\noindent \textbf{Autoregressive Modeling with Transformer.} After obtaining the sequence that contains the necessary image features at step $t$, we feed it into the transformer and predict the feature index $\hat{q_{l_t}}$ for current target location $l_t$. We then fill current target location $l_t$ with the latent feature corresponding to the predicted feature index $\hat{q_{l_t}}$. We learn this model by maximizing the log-likelihood of the latent representations and perform this process (including conditional cross-attention step) autoregressively until all locations are filled. 

\subsection{Training and Inference}
\noindent \textbf{Training.} During training, we randomly select one view from the rendered images as input and sample one shape from the corresponding mappings as the ground-truth. We employ teacher-forcing by setting filled locations as ground-truth features during autoregressive learning. We also fine-tune the pretrained CLIP encoder to make it more adaptable to our task. The loss function is the cross entropy with respect to feature indexes for each location.

\noindent \textbf{Inference.} During inference, we complete an empty query sequence conditioned on the image in an autoregressive manner from the transformer. At step $t$, we have:
\begin{equation}
    p_{l_t}(e') = T_{\theta}(\left\{ \hat{e'_{l_{t-1}}}, \hat{e'_{l_{t-2}}}, \hat{e'_{l_{t-3}}}, \cdots  \right\}, l_{t}, c).
\end{equation}
where $l_{t}$ represents current location and $\hat{e'_{l_t}}$ is generated by sampling from the distribution $p_{l_t}(e')$. We then feed the complete sequence into $D_{\phi}$ to generate one plausible 3D shape.

\noindent \textbf{Implementation Details.} To train our network, we use ADAM as the optimizer with an initial learning rate of 1e-5 for the CLIP image encoder and 1e-4 for other components. We decay both learning rates by multiplying 0.9 every 10,000 iterations. The batch size is fixed to 10 and it takes around 72 hours to converge on a single GTX 1080Ti GPU. 
\section{Multi-Hypothesis Data Augmentation}
\label{sec:data_generation}
As discussed in Sec.~\ref{sec:introduction}, we intend to create simulated image-to-shape training pairs that further enable improved fine-tuning for real-world scenarios.  To this end, we render CAD models from ShapeNetCore~\cite{chang2015shapenet} in tailored settings such that objects are captured under strong occlusion or truncation. We focus on common indoor furniture models including $chair$, $table$, $cabinet$, $bed$, and $bookshelf$. Due to inherent ambiguities in our images, we also establish ground-truth mappings using part-level annotations from PartNet~\cite{mo2019partnet}, which allows for multiple distinct ground-truth shapes per image 

\subsection{Rendering CAD Models with Ambiguity}
\begin{figure}[htbp]
\centering
\begin{tabular}{ccc}
\bmvaHangBox{\fbox{\includegraphics[width=0.2\textwidth]{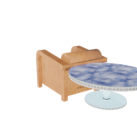}}}&
\bmvaHangBox{\fbox{\includegraphics[width=0.2\textwidth]{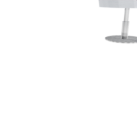}}}&
\bmvaHangBox{\fbox{\includegraphics[width=0.2\textwidth]{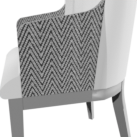}}}\\
(a)  & (b) & (c)
\end{tabular}
\caption{\textbf{Different Types of Ambiguity.} (a) indicates ambiguity caused by occlusion. (b) and (c) represent ambiguity resulted from truncation via limited field-of-view of the camera.
}
\label{fig:ambiguity_examples}
\end{figure}

\noindent \textbf{Occlusion.} To induce occlusion, we first randomly rotate the target object at the origin along the upward axis to capture varying levels of self-occlusion. Then, we randomly select another object to act as the occluding object, positioned between the virtual camera and the target object. This setup ensures two conditions: (1) no overlap between the occluding and target objects in 3D space, and (2) reasonable occlusion of the target object by the occluding one in the 2D image. Fig.~\ref{fig:ambiguity_examples}(a) provides an example.

\noindent \textbf{Truncation.}
Truncation typically occurs when parts of an object are obscured by the limited field-of-view of the camera. We consider two scenarios: First, we move the object away from the center of the image plane as shown in Fig.~\ref{fig:ambiguity_examples}(b). Second, we simulate scenarios where the camera is too close to the object, as displayed in Fig.~\ref{fig:ambiguity_examples}(c).

\subsection{Creating Multi-Hypothesis Image-to-Shape Pairs}
The inherent ambiguity in our renderings allows for the coexistence of multiple viable 3D ground-truth shapes, each corresponding to a plausible interpretation of the image's content. Therefore, we take this fact into account to create our image-to-shape training pairs. We first classify available CAD models into similar groups. Then we construct the ground-truth mappings for each image from these groups by considering the geometry of visible parts and per-pixel part labels. Further details can be found in the supplementary material. Fig.~\ref{fig:sample_mapping} shows one sample mapping. On average, each rendering maps to two ground-truth shapes accurately, with $l_2$ chamfer distance of visible parts below 0.04.
\begin{figure}[ht]
\centering
\begin{tabular}{cccc}
\bmvaHangBox{\fbox{\includegraphics[width=0.19\textwidth]{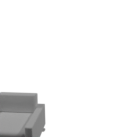}}}&
\bmvaHangBox{\fbox{\includegraphics[width=0.19\textwidth]{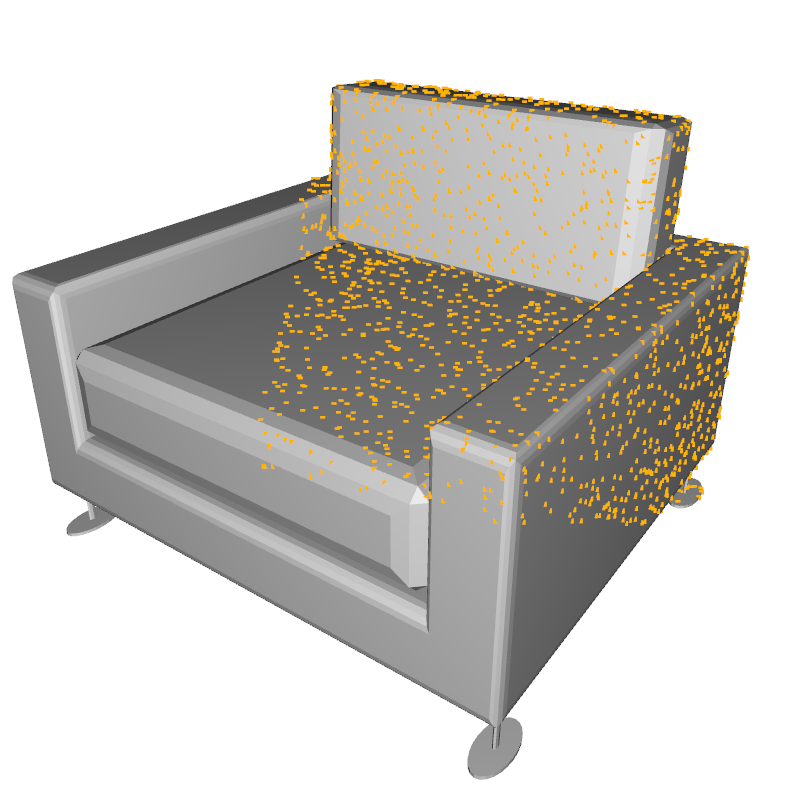}}}&
\bmvaHangBox{\fbox{\includegraphics[width=0.19\textwidth]{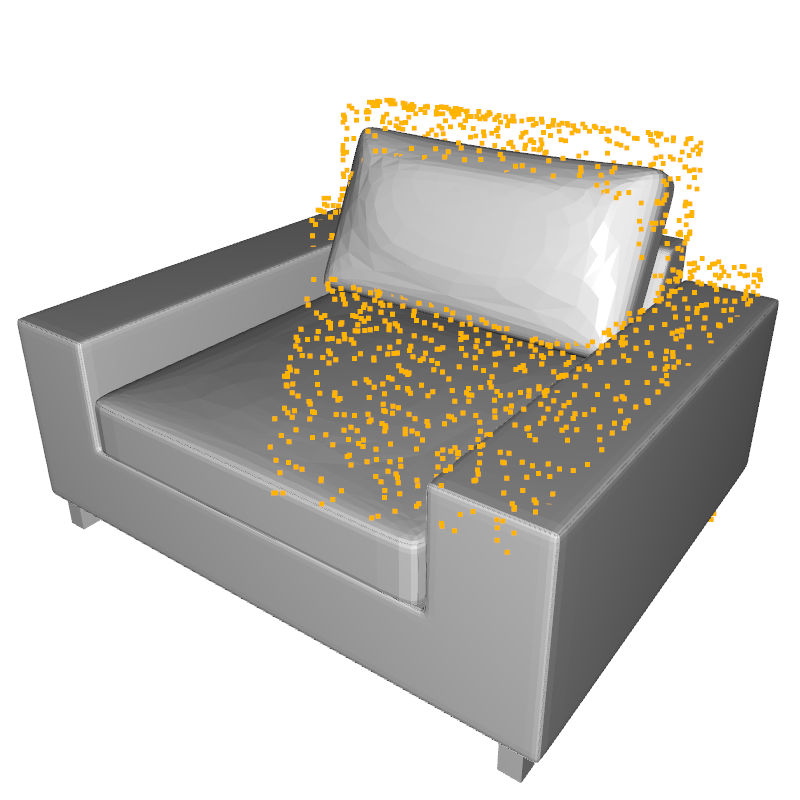}}}&
\bmvaHangBox{\fbox{\includegraphics[width=0.19\textwidth]{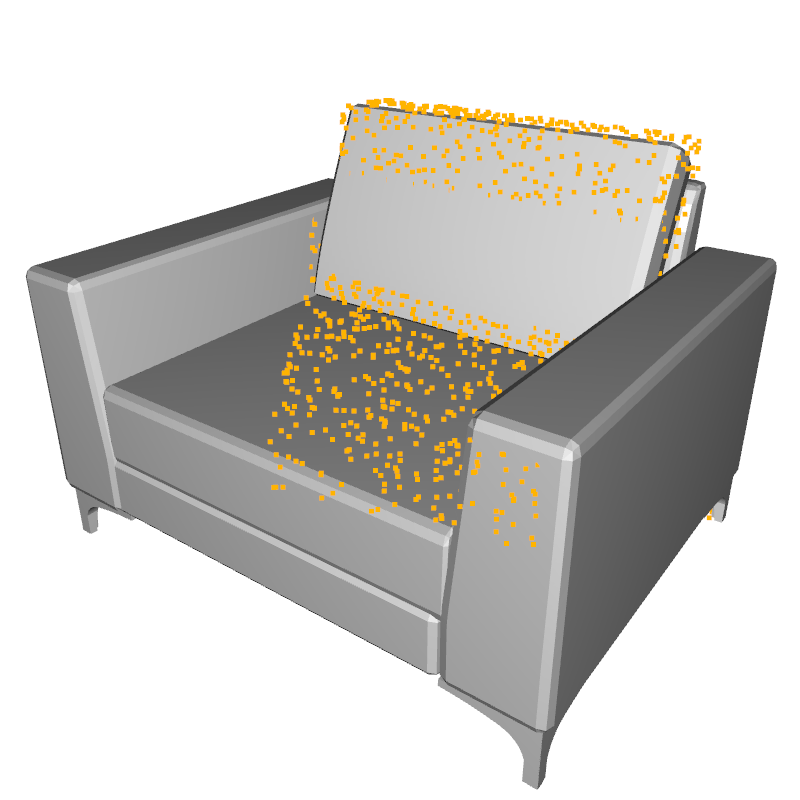}}}\\
(a) Rendering & (b) 1st ground-truth & (c) 2nd ground-truth & (d) 3rd ground-truth
\end{tabular}
\caption{\textbf{Ground-Truth Mapping.} (a) represents a sample rendering and (b), (c) and (d) represent multiple distinct ground-truth shapes that corresponds to (a). Yellow points are for visualization purposes only, indicating the overlap between ground-truth visible parts and multiple plausible shapes.
}
\label{fig:sample_mapping}
\end{figure}
\section{Experiments}
\label{sec:experiments}
We first train and test our model on our synthetic dataset following splits provided by ShapeNet~\cite{chang2015shapenet}. We then fine-tune the CLIP image encoder and conditional cross-attention module of our pretrained model on real-world data from ScanNet~\cite{dai2017scannet,avetisyan2019scan2cad}. We use the ScanNet25k image data and use ground-truth semantic instance masks to generate per-instance images for training without background. While for validation set, we adopt machine-generated masks produced by ODISE~\cite{xu2023open}. Then we align each instance image with ShapeNet CAD models using annotations provided by Scan2CAD~\cite{avetisyan2019scan2cad}. We split the generated images according to provided train/validation scenes.
\subsection{Evaluation Metrics}
Our aim is to generate shapes that capture the distribution of possible reconstructions that explain an input image observation.
Thus, we evaluate both diversity of reconstructed shapes as well as their reconstruction quality.
To assess the diversity in generated shapes, we employ the Total Mutual Difference (TMD), computed as the average chamfer distance among $k$ generated shapes. We set $k=6$ in our experiments, aligning with the average number of models within a similar group (refer to Sec.~\ref{sec:data_generation}). For evaluating the generation quality, we utilize bidirectional $l_2$ chamfer distance (CD) and F-score@1\%~\cite{tatarchenko2019singleview} as metrics. We present the minimal CD and the highest F-score among the $k$ generated shapes in our results.

\subsection{Comparison to State of the Art}
\begin{table}[htbp]
\Huge
\renewcommand\arraystretch{1.5}
    \centering
    \resizebox{\textwidth}{!}{
    \begin{tabular}{|c|ccc|ccc|ccc|ccc|ccc|ccc|c}
    \hline
     & \multicolumn{3}{c|}{Chair} & \multicolumn{3}{c|}{Table} & \multicolumn{3}{c|}{Cabinet} & \multicolumn{3}{c|}{Bed} & \multicolumn{3}{c|}{Bookshelf} & \multicolumn{3}{c|}{Average}\\
    \cline{2-19}
       & TMD $\uparrow$& CD $\downarrow$ & F-score $\uparrow$ & TMD $\uparrow$& CD $\downarrow$ & F-score $\uparrow$ & TMD $\uparrow$& CD $\downarrow$ & F-score $\uparrow$ & TMD $\uparrow$& CD $\downarrow$ & F-score $\uparrow$ & TMD $\uparrow$& CD $\downarrow$ & F-score $\uparrow$ & TMD $\uparrow$& CD $\downarrow$ & F-score $\uparrow$\\
      \hline
      Pix2Vox~\cite{xie2019pix2vox} & - & 5.99 & 0.175 & - & 7.03 & 0.164 & - & 
6.52 & 0.114 & - & 11.8 & 0.115 & - & 9.85 & 0.125 & - & 8.24 & 0.139 \\
      AutoSDF~\cite{mittal2022autosdf} & 0.041 & 3.19 & 0.250 & 0.040 & 3.81 & 0.280 & 0.047 & 4.70 & 0.160 & 0.054 & 8.78 & 0.135 & 0.042 & 4.47 & 0.247 & 0.045 & 5.04 & 0.215\\
      SDFusion~\cite{cheng2023sdfusion} & \textbf{0.073} & 3.32 & 0.233 & \textbf{0.078} & 3.93 & 0.249 & \textbf{0.103} & 4.77 & 0.145 &0.109 & \textbf{5.89} & 0.119 & \textbf{0.093} & 4.69 & 0.226 & \textbf{0.091} &4.52 &0.194\\
      Ours & 0.052 & \textbf{2.89} & \textbf{0.260} & 0.058 & \textbf{3.55} & \textbf{0.284} & 0.066 & \textbf{3.76} & \textbf{0.184} & 0.075 & 7.54 & \textbf{0.141} & 0.061 & \textbf{4.31} & \textbf{0.254} & 0.062 & \textbf{4.41} & \textbf{0.224}\\
      \cline{2-19}
      \hline
    \end{tabular}
    }
    \caption{\textbf{Results Comparisons on Synthetic Data}. Our method generates higher quality shapes compared with other baselines in terms of CD and F-score, with reasonable diversity. 
    }
    \label{table:synthetic_results}
    \vspace{-0.3cm}
\end{table}

\begin{table}[htbp]
    \centering
	\resizebox{0.4\linewidth}{!}{
    \begin{tabular}{|c|c|c|}
    \hline
    & ShapeNet~\cite{chang2015shapenet} & ScanNet~\cite{dai2017scannet,avetisyan2019scan2cad}\\
    \hline
    SDFusion~\cite{cheng2023sdfusion} & 6.71 & 10.4 \\
    \hline
    Ours & \textbf{5.41} & \textbf{5.89}\\
    \hline 
    \end{tabular}
    }
    \caption{\textbf{
    Average One-Way $l_2$ Chamfer Distance between Visible Ground-Truth Points with Generated Hypotheses.} Our method generates shapes that align better with the input image on both synthetic and real-world dataset.
    }
    \label{table:visible_cd}
\end{table}
\noindent \textbf{Results on Synthetic Data.} Tab.~\ref{table:synthetic_results} compares our method with state-of-the-art single-image based methods Pix2Vox~\cite{xie2019pix2vox}, AutoSDF~\cite{mittal2022autosdf} and SDFusion~\cite{cheng2023sdfusion} on synthetic dataset. Pix2Vox is fully deterministic, and so only provides one hypothesis. Our method is able to generate higher quality shapes compared with other baselines. In terms of generation diversity, SDFusion is capable of generating more diverse shapes. However, the higher diversity comes at the cost of outputs that do not align well with the input data. To validate this, we calculate the average one-way $l_2$ chamfer distance between visible ground-truth points and generated hypotheses. The results, presented in Tab.~\ref{table:visible_cd}, illustrate that our model produces shapes with better alignment compared to those generated by SDFusion. This suggests the lacking of plausibility among higher diverse shapes produced by SDFusion, thereby demonstrating more reasonable diversity among hypotheses produced by our method. 

\begin{figure}[ht]
    \centering
    \includegraphics[width=1.0\linewidth]{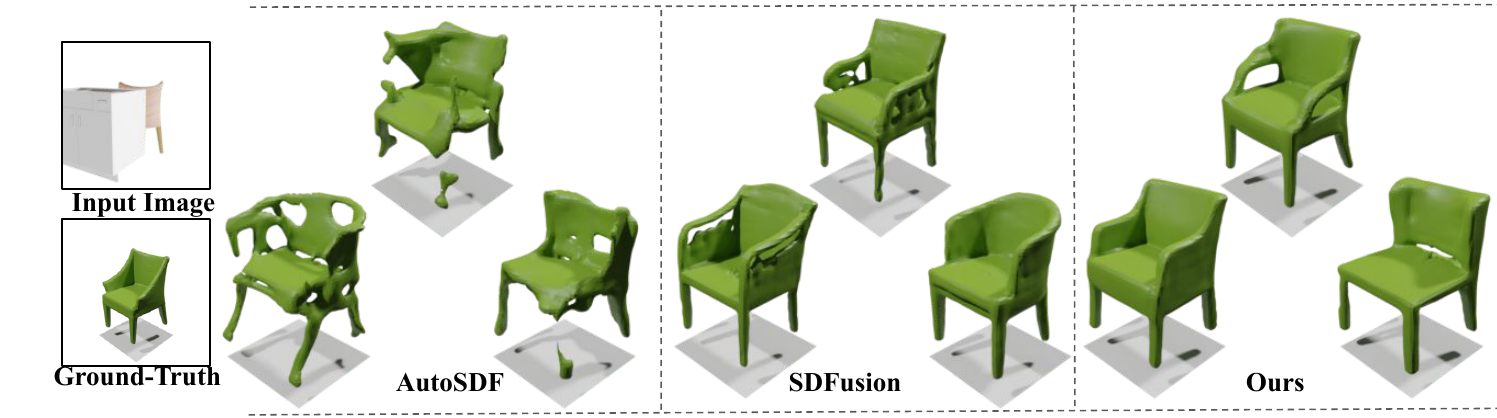}
    \includegraphics[width=1.0\linewidth]{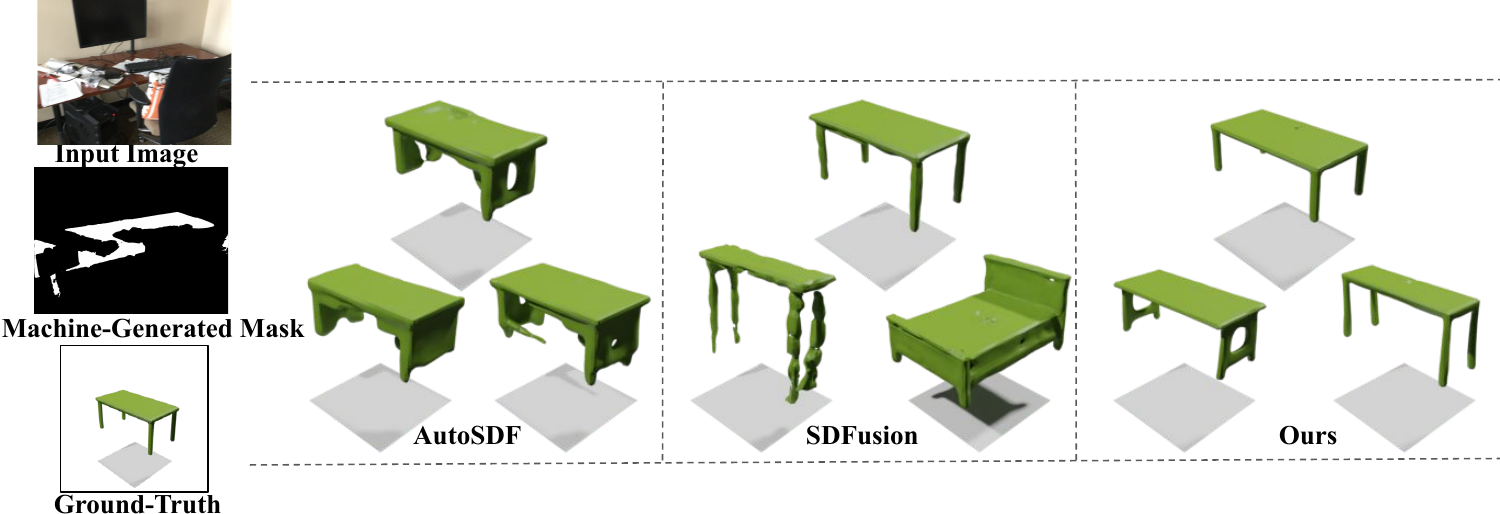}
    \caption{\textbf{Qualitative Comparisons.} Top: results on synthetic images; bottom: results on real images. Our method generates higher quality and more plausible hypotheses.
    }
    \label{fig:viz}
\end{figure}

In Fig.\ref{fig:viz} top part, we present qualitative comparisons of synthetic data among AutoSDF, SDFusion and our method. The visual inspection clearly demonstrates that our method excels in generating higher-fidelity shapes from a single highly ambiguous RGB image compared to other baselines, with a more reasonable diversity level.

\begin{table}[htbp]
\Huge
\renewcommand\arraystretch{1.5}
    \centering
    \resizebox{\textwidth}{!}{
    \begin{tabular}{|c|ccc|ccc|ccc|ccc|ccc|ccc|c}
    \hline
     & \multicolumn{3}{c|}{Chair} & \multicolumn{3}{c|}{Table} & \multicolumn{3}{c|}{Cabinet} & \multicolumn{3}{c|}{Bed} & \multicolumn{3}{c|}{Bookshelf} & \multicolumn{3}{c|}{Average}\\
    \cline{2-19}
       & TMD $\uparrow$ & CD $\downarrow$ & F-score $\uparrow$ & TMD $\uparrow$ & CD $\downarrow$ & F-score $\uparrow$ & TMD $\uparrow$& CD $\downarrow$ & F-score $\uparrow$ & TMD $\uparrow$& CD $\downarrow$ & F-score $\uparrow$ & TMD $\uparrow$& CD $\downarrow$ & F-score $\uparrow$ & TMD $\uparrow$& CD $\downarrow$ & F-score $\uparrow$\\
      \hline
      Pix2Vox~\cite{xie2019pix2vox} & - & 7.01 & 0.158 & - & 11.1 & 0.157 & - & 12.8 & 0.080 & - & 10.2 & 0.161 & - & 7.14 & 0.119 & - & 9.68 & 0.135 \\
      AutoSDF~\cite{mittal2022autosdf} & 0.019 & 5.33 & 0.174 & 0.019 & 9.46 & 0.169 & 0.030 & 10.9 & 0.113 & 0.019 & 8.24 & 0.196 & 0.020 & 7.59 & 0.170 & 0.021 & 8.32 & 0.164\\
      SDFusion~\cite{cheng2023sdfusion} & \textbf{0.104} & 4.52 & 0.197 & \textbf{0.152} & 5.66 & 0.207 & \textbf{0.141} & 6.37 & 0.166 & \textbf{0.135} & 4.54 & 0.216 & \textbf{0.129} & 3.75 & 0.185 & \textbf{0.132} & 4.97 & 0.200\\
      Ours & 0.077 & \textbf{3.69} & \textbf{0.222} & 0.096 & \textbf{3.64} & \textbf{0.234} & 0.130 & \textbf{4.39} & \textbf{0.183} & 0.096 & \textbf{3.17} & \textbf{0.319} & 0.081 & \textbf{3.18} & \textbf{0.207} & 0.096 & \textbf{3.61} & \textbf{0.228}\\
      \cline{2-19}
      \hline
    \end{tabular}
    }
    \caption{\textbf{Results Comparisons on Real-World Data}. Our method generates higher quality shapes compared with other baselines in terms of CD and F-score, with reasonable diversity.}
    \label{table:real-world_results}
\end{table}

\noindent \textbf{Results on Real-World Data.} We then compare the experimental results obtained from ScanNet~\cite{dai2017scannet,avetisyan2019scan2cad} images. As shown in Tab.~\ref{table:real-world_results}, our model surpasses state of the art in terms of generation quality. Additionally, the results in Tab.~\ref{table:visible_cd} indicate that SDFusion~\cite{cheng2023sdfusion} produces nearly twice the visible chamfer distance compared to ours. This further emphasizes our model's capability to generate shapes with more reasonable diversity.

In Fig.\ref{fig:viz} bottom part, we present qualitative comparisons of real-world data among AutoSDF, SDFusion and our method. Our method achieves higher shape generation quality compared to other baselines. Despite the higher diversity among hypotheses produced by SDFusion, they lack alignment with the input image.

\subsection{Ablation Studies}
\begin{table}[htbp]
    \centering
	\resizebox{0.45\linewidth}{!}{
    \begin{tabular}{|c|c|c|c|}
    \hline
    & TMD $\uparrow$& CD $\downarrow$ & F-score $\uparrow$\\
    \hline
    w/o mapping & 0.060 & \textbf{4.26} & \textbf{0.230}\\
    w/ coarse mapping & 0.066 & 4.43 & 0.216 \\
    \hline 
    average pooling & 0.065 & 4.53 & 0.221\\
    max pooling & 0.066 & 4.50 & 0.219\\
    \hline
    w/o concatenation & \textbf{0.105} & 39.1 & 0.032\\
    \hline
    Ours & 0.062 & 4.41 & 0.224\\
    \hline
    \end{tabular}
    }
    \caption{\textbf{Ablations on Synthetic Data.} Our model generates more diverse shapes being trained with multiple ground-truth mappings. Cross-attention is effective in generating higher quality shapes compared with average pooling and max pooling. Without final concatenation, very little is learned by the model.}
    \label{table:ablation_synthetic_dataset}
    \vspace{-0.3cm}
\end{table}

\begin{table}[htbp]
    \centering
	\resizebox{0.35\linewidth}{!}{
    \begin{tabular}{|c|c|c|c|}
    \hline
    & TMD$\uparrow$ & CD $\downarrow$ & F-score $\uparrow$\\
    \hline
    from scratch & 0.044 & 6.87 & 0.203 \\
    \hline
    Ours & \textbf{0.096} &\textbf{3.61} & \textbf{0.228} \\
    \hline 
    \end{tabular}
    }
    \caption{\textbf{Ablations on Real-World Data.} Our multi-hypothesis data augmentation with synthetic data enables pre-training that notably improves shape reconstruction results when fine-tuned on real-world data.}
    \label{table:ablation_real-world_dataset}
\end{table}

\noindent \textbf{Does cross-attention help?} In our model, we employ cross-attention as the image conditional mechanism, indicating a weighted sum of image features. We compare it with two similar strategies: (1) average pooling, where all weights are the same, and (2) max pooling, where only one non-zero weight is used. Results in Tab.~\ref{table:ablation_synthetic_dataset} demonstrate that our model with cross-attention generates higher-quality shapes compared with two other strategies.

\noindent \textbf{Does concatenation help?} Within our conditional cross-attention module, the raw cross-attention outputs merely contain image features. We thus concatenate the original 3D embeddings with them in the final step. To analyze its effect, we conduct an additional experiment by removing the concatenation step, using raw cross-attention outputs for subsequent operations. Results in Tab.~\ref{table:ablation_real-world_dataset} reveal that the model without concatenation learns very little. 

\noindent \textbf{Does one image mapping to multiple ground-truths improve generation diversity?} To analyze our mapping strategy, we conduct two additional experiments: training our model (1) without mapping and (2) with coarse mapping by setting all shapes within a similar group as ground-truths. Results in Tab.~\ref{table:ablation_synthetic_dataset} show that our method trained with mapping achieves greater diversity in generated shapes compared to the version without mapping. Furthermore, opting for coarse mapping further increases diversity, demonstrating our method's ability to capture the varying diversity of ground-truth data.

\noindent \textbf{Does pretraining on the synthetic dataset improve real-world reconstruction results?} To assess the efficacy of our synthetic dataset, we conduct experiments by training our model from scratch on real-world data and comparing the reconstruction results with our fine-tuned version. Results from Tab.~\ref{table:ablation_real-world_dataset} demonstrate the superior performance of our fine-tuned version, thereby affirming the effectiveness of our synthetic dataset.

\subsection{Limitations} In our image-based shape generation framework, we utilize patch-wise image encodings as  conditional input, enabling  robust global alignment between input object observations and output reconstructions. However, this may not fully capture the shape variance and ambiguities that lie in the local details, such as the wheels of the chair present in Fig.~\ref{fig:sample_mapping}. A potential approach to overcome this limitation involves adopting pixel-wise image features that are more locally extracted. In addition, while our method achieves promising reconstruction results at object-level, generating scene-level reconstructions could provide richer perception for applications like robotics. In our future work, we consider expanding the generation scale from object-level to scene-level, leveraging monocular views.
\section{Conclusion}
\label{sec:conclusion_and_future_work}
In this work, we propose a new approach for generating the probabilistic distribution of 3D shape reconstructions conditioned on a highly ambiguous RGB image. We employ cross-attention to effectively identify the most relevant region of interest from the input image for shape generation. We also introduce a synthetic data augmentation approach to enable effective multi-hypothesis generation, by computing  multiple potential ground-truth shapes that correspond to a single image observation during training. This synthetic augmentation enables pre-training that notably improves shape reconstruction results when fine-tuned on real-world data. Experiments demonstrate that our model outperforms state of the art in both generation quality and plausible generation diversity. We hope our work will draw increased attention to this important and realistic problem, highlighting the need for innovative solutions in 3D shape reconstruction from single ambiguous image data.

\section{Acknowledgements}
This work was supported by the ERC Starting Grant SpatialSem (101076253).

\bibliography{egbib}
\clearpage
\section{Supplementary Material}
In Sec.~\ref{subsec:additional_results}, we present more qualitative comparisons on both synthetic and real-world data. We also provide more implementation details in Sec.~\ref{subsec:further_details}.
\subsection{Additional Qualitative Comparisons}
\label{subsec:additional_results}
\begin{figure}[htbp]
    \centering
    \includegraphics[width=1.0\linewidth]{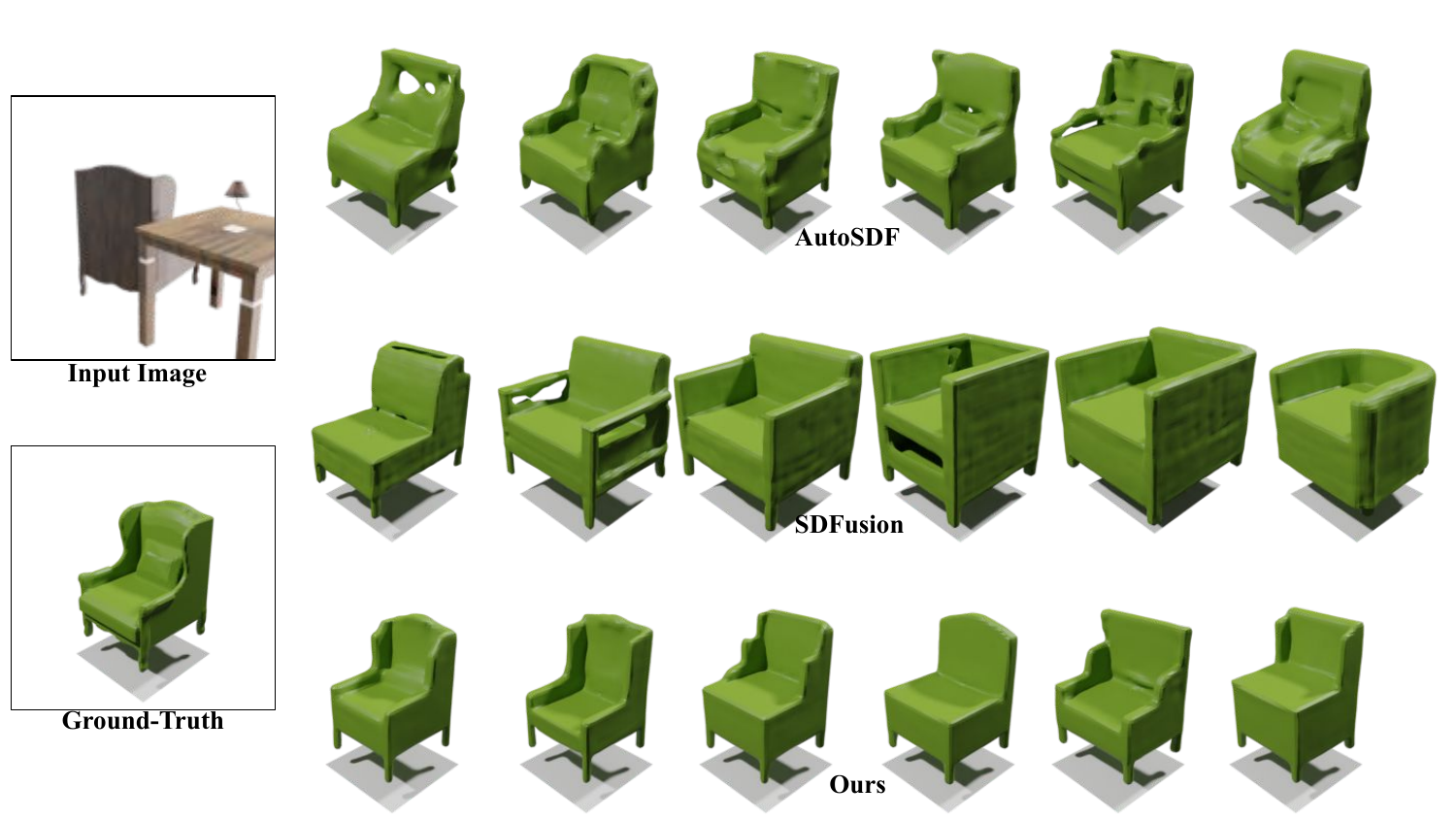}
    \hrule
    \includegraphics[width=1.0\linewidth]{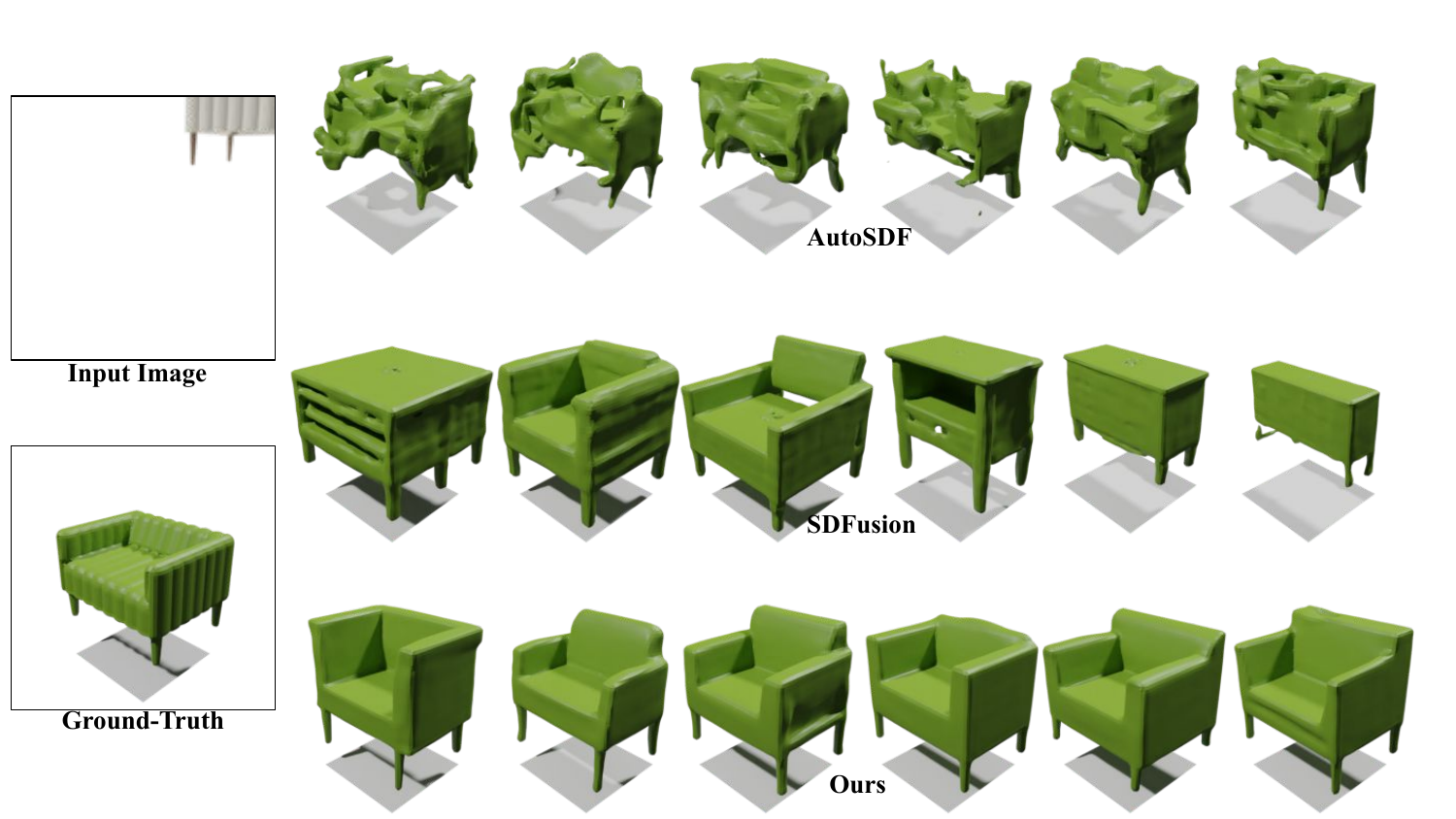}
    \caption{\textbf{More Qualitative Comparisons on Synthetic Data.} Our method generates higher quality and more plausible hypotheses compared with other baselines.}
    \label{fig:viz_supp_synthetic_1}
\end{figure}

\begin{figure}[htbp]
    \centering
    \includegraphics[width=1.0\linewidth]{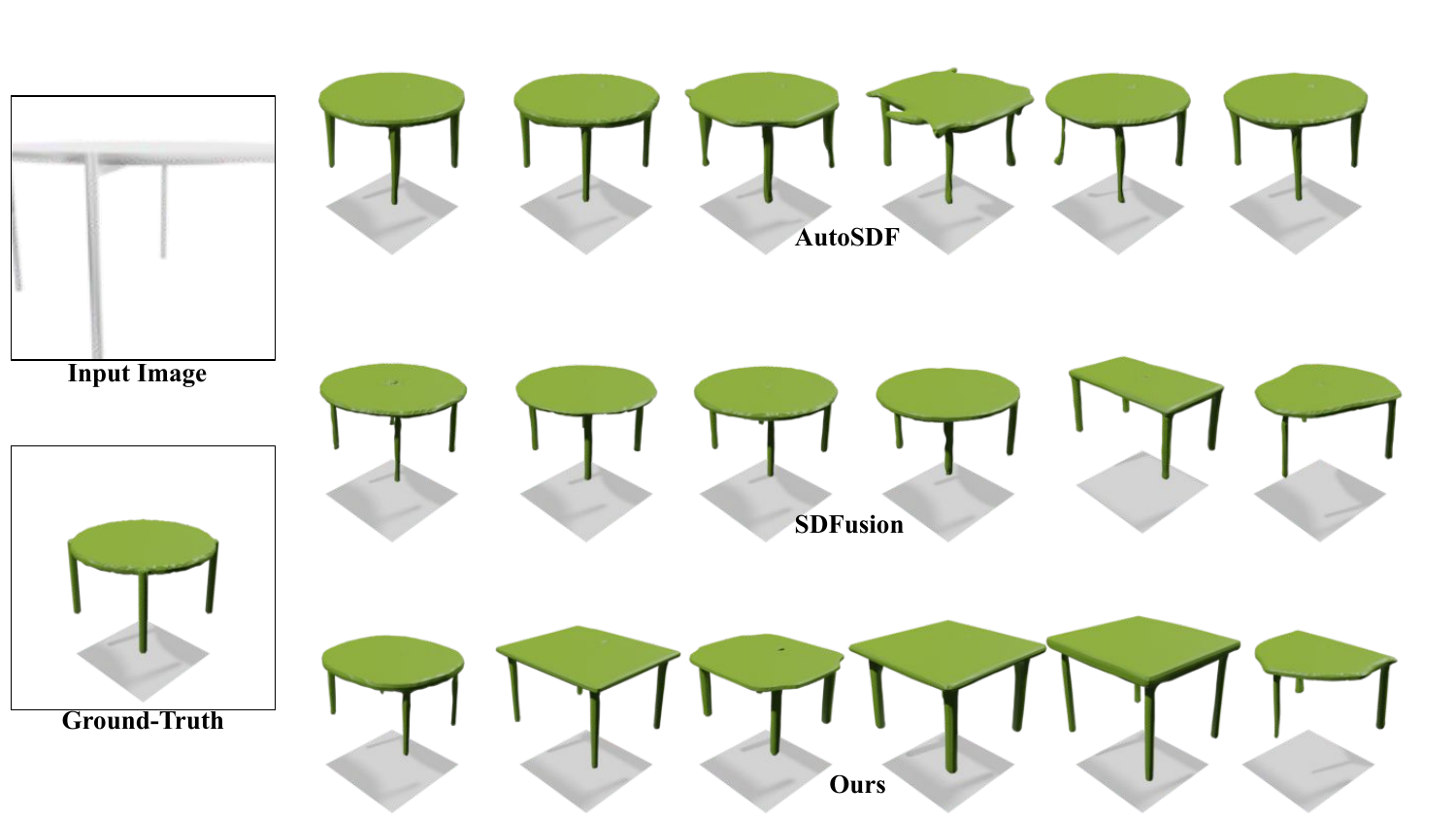}
    \hrule
    \includegraphics[width=1.0\linewidth]{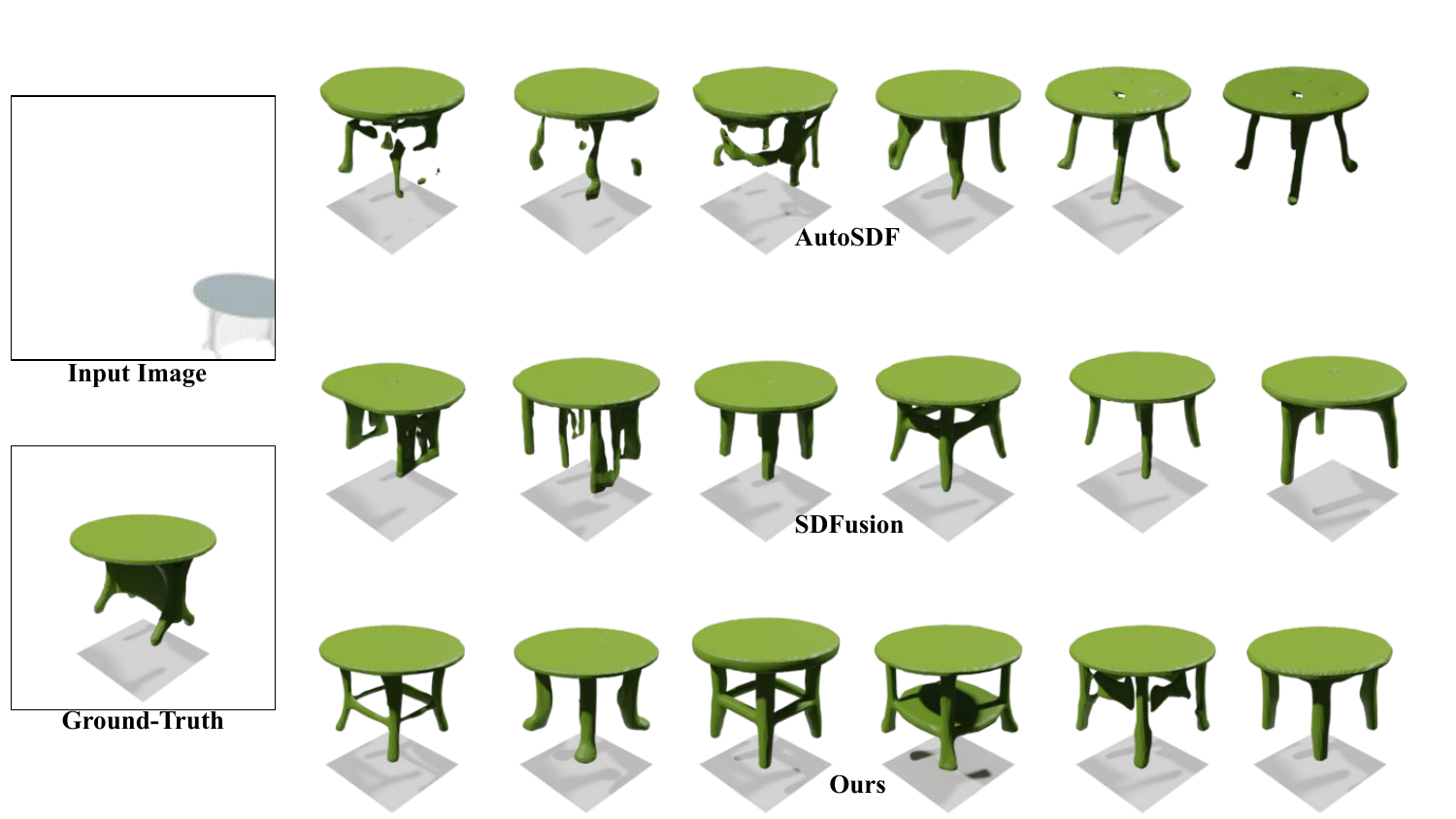}
    \caption{\textbf{More Qualitative Comparisons on Synthetic Data.} Our method generates higher quality and more plausible hypotheses compared with other baselines.}
    \label{fig:viz_supp_synthetic_2}
\end{figure}

\begin{figure}[htbp]
    \centering
    \includegraphics[width=1.0\linewidth]{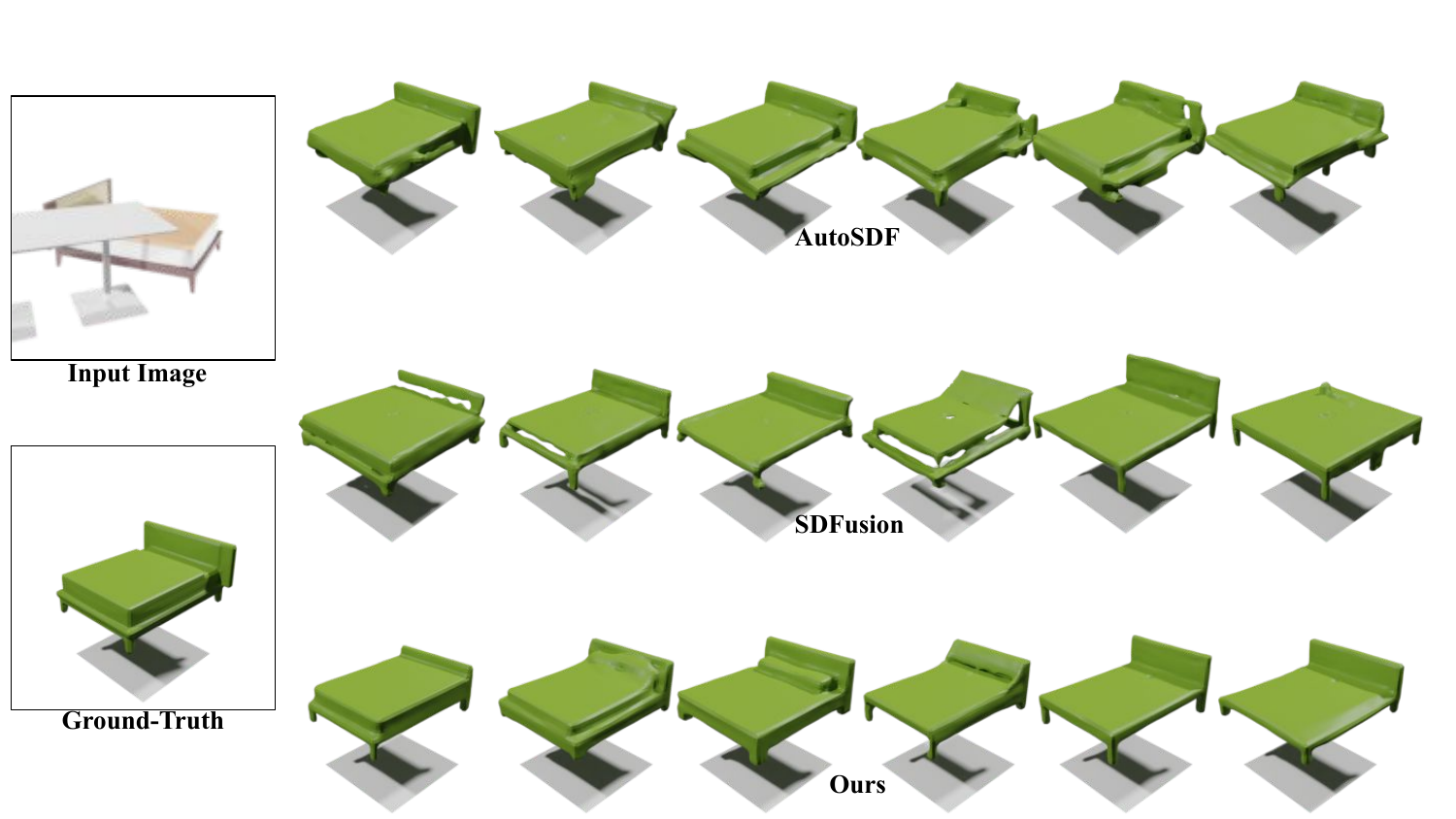}
    \hrule
    \includegraphics[width=1.0\linewidth]{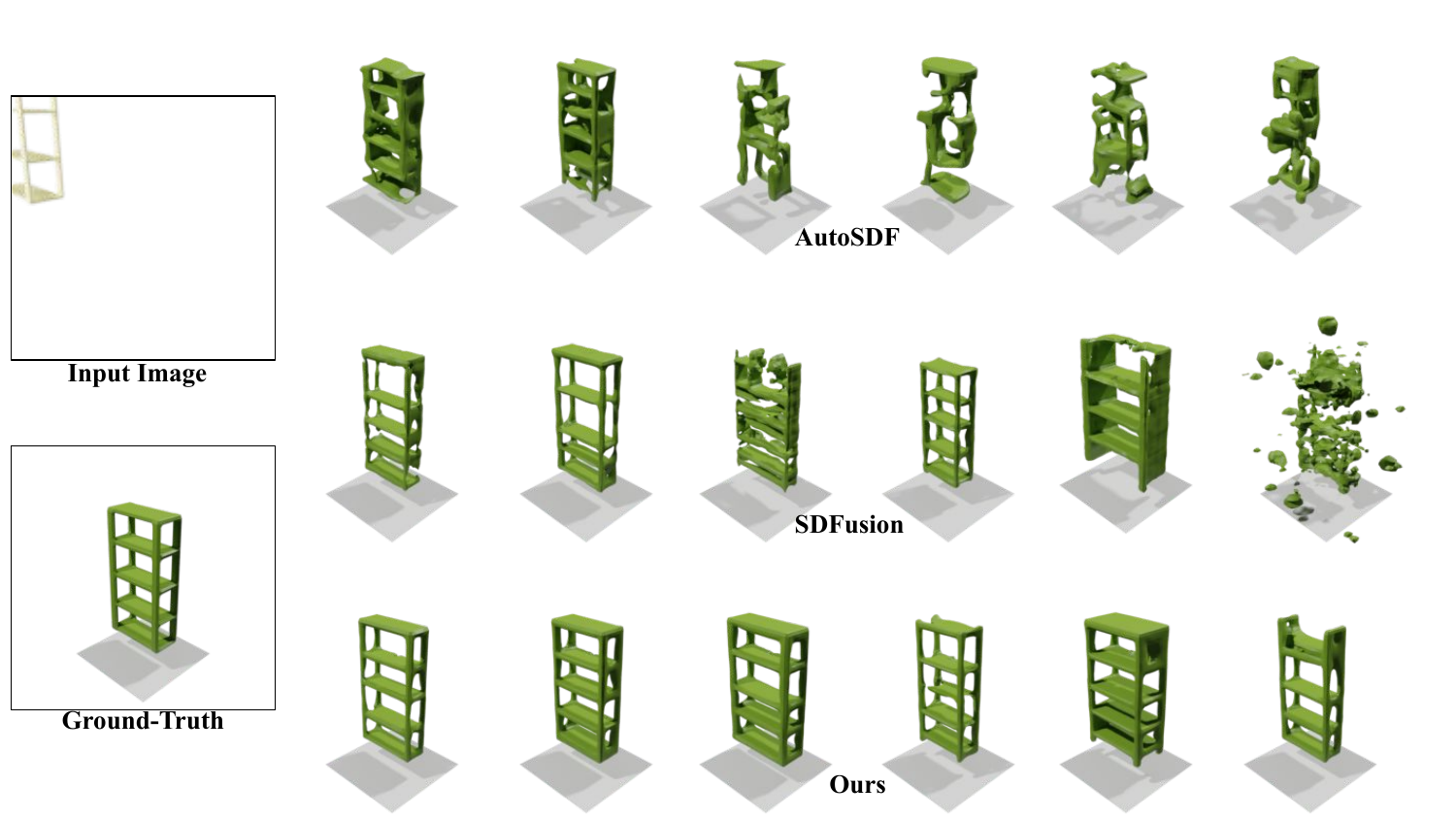}
    \caption{\textbf{More Qualitative Comparisons on Synthetic Data.} Our method generates higher quality and more plausible hypotheses compared with other baselines.}
    \label{fig:viz_supp_synthetic_3}
\end{figure}

\begin{figure}[htbp]
    \centering
    \includegraphics[width=1.0\linewidth]{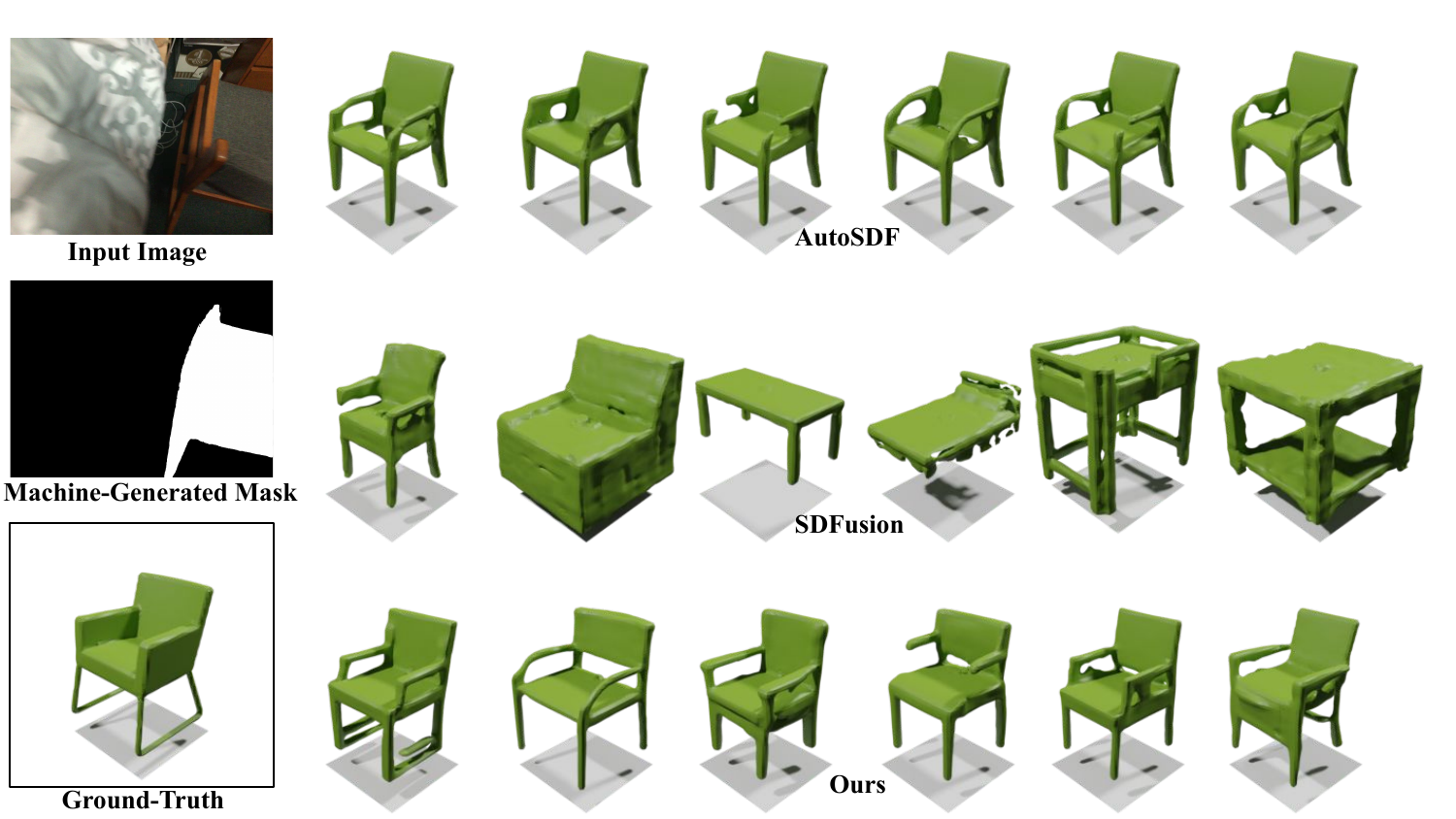}
    \hrule
    \includegraphics[width=1.0\linewidth]{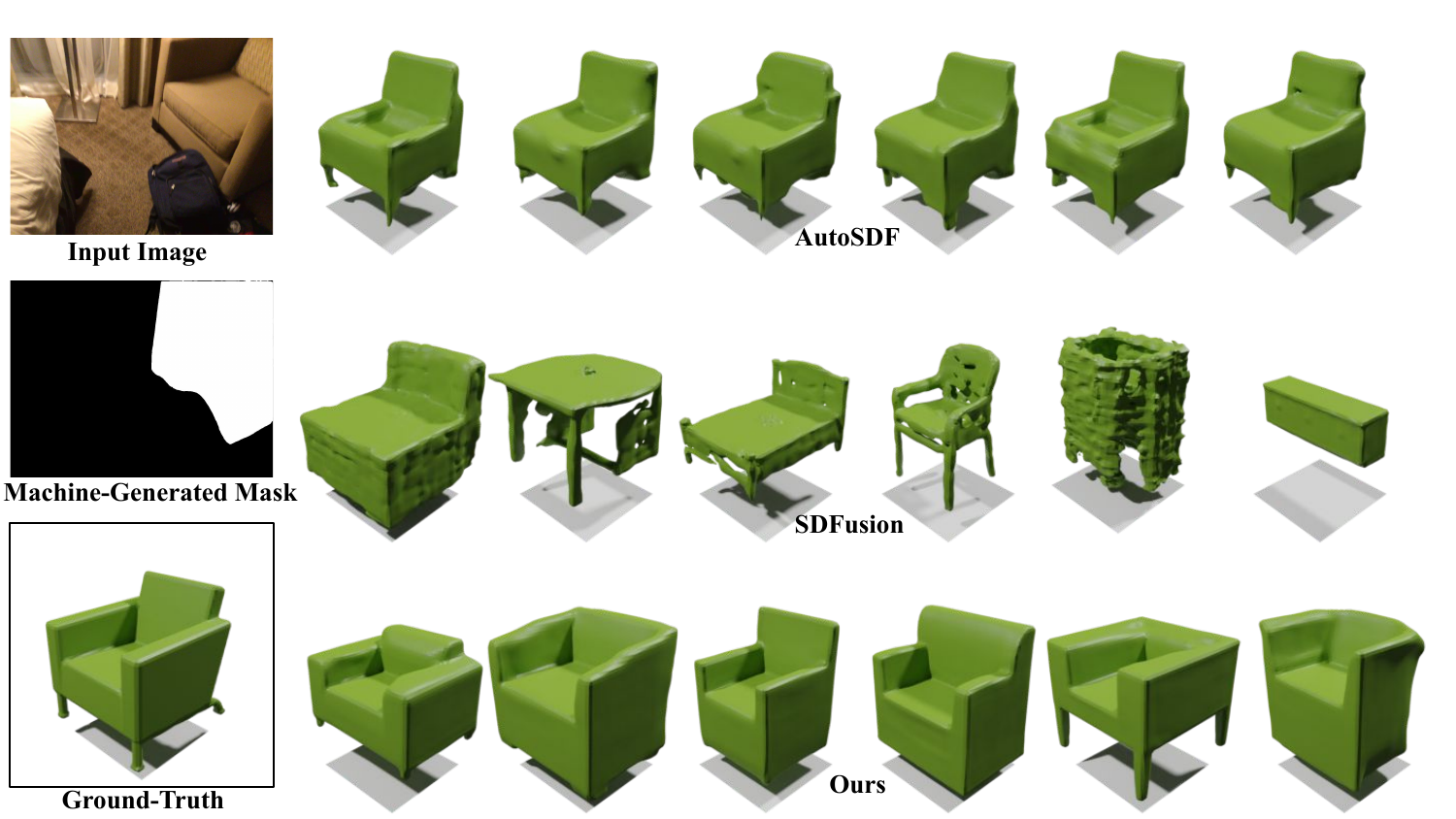}
    \caption{\textbf{More Qualitative Comparisons on Real-World Data.} Our method generates higher quality and more plausible hypotheses compared with other baselines.}
    \label{fig:viz_supp_real-world_1}
\end{figure}

\begin{figure}[htbp]
    \centering
    \includegraphics[width=1.0\linewidth]{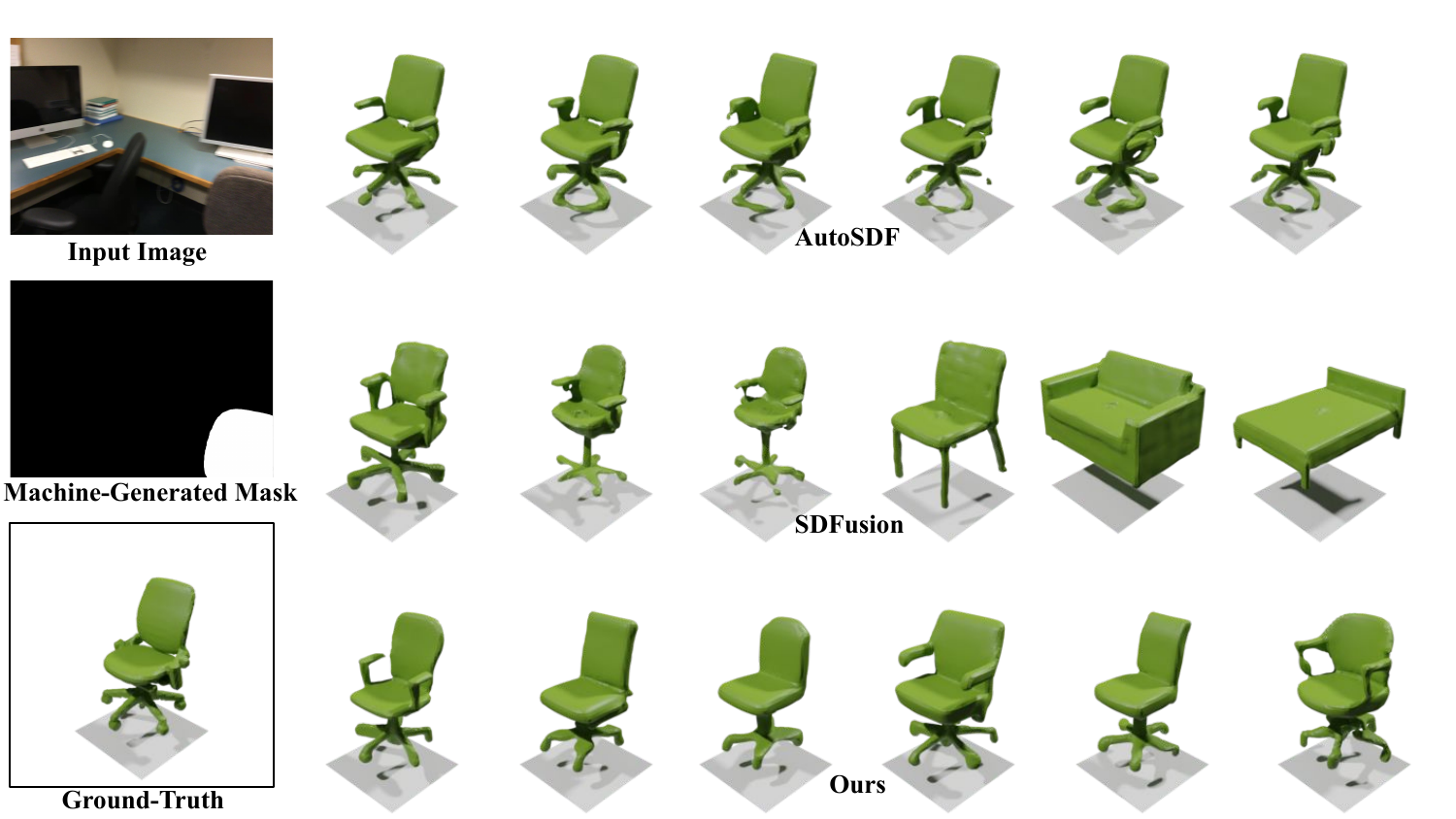}
    \hrule
    \includegraphics[width=1.0\linewidth]{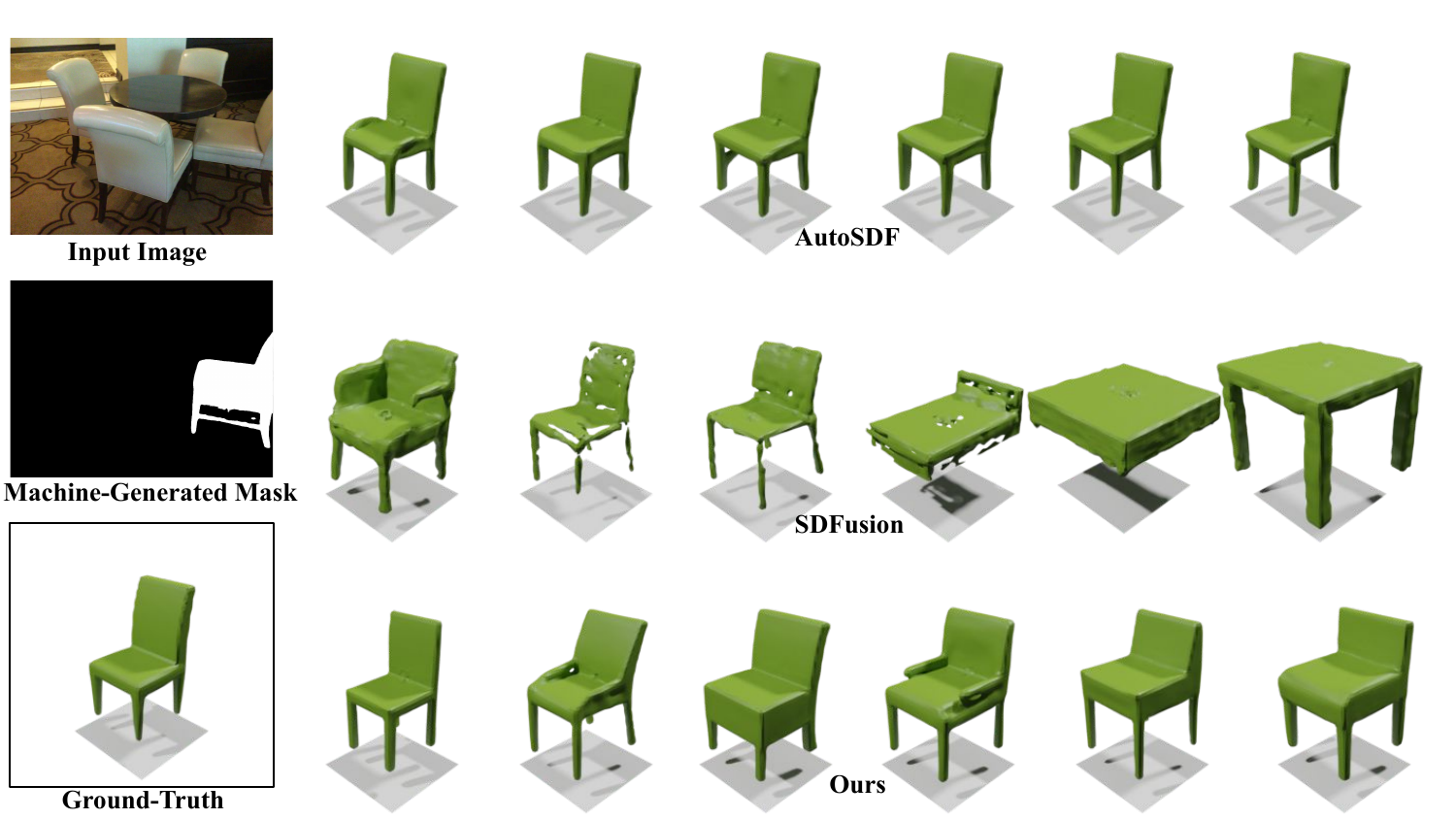}
    \caption{\textbf{More Qualitative Comparisons on Real-World Data.} Our method generates higher quality and more plausible hypotheses compared with other baselines.}
    \label{fig:viz_supp_real-world_2}
\end{figure}

We present more qualitative comparisons among AutoSDF, SDFusion and our method. Fig.~\ref{fig:viz_supp_synthetic_1}, Fig.~\ref{fig:viz_supp_synthetic_2} and Fig.~\ref{fig:viz_supp_synthetic_3} contain visualizations from synthetic data. Fig.~\ref{fig:viz_supp_real-world_1}, Fig.~\ref{fig:viz_supp_real-world_2} and Fig.~\ref{fig:viz_supp_real-world_3} provides qualitative samples from real-world data. Our method generally generates higher quality and more plausible hypothesis shapes compared with other baselines.

\begin{figure}[htbp]
    \centering
    \includegraphics[width=1.0\linewidth]{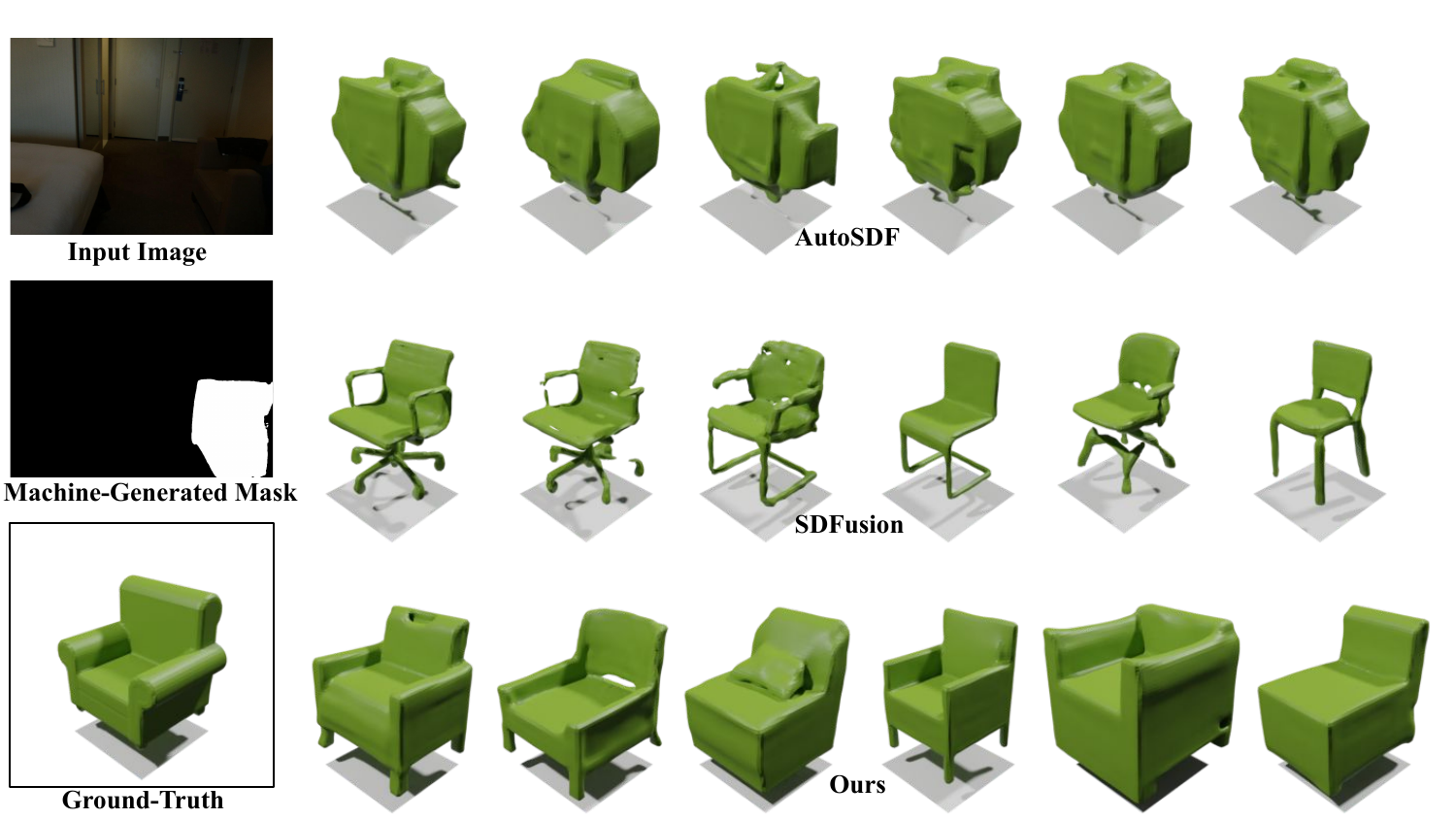}
    \hrule
    \includegraphics[width=1.0\linewidth]{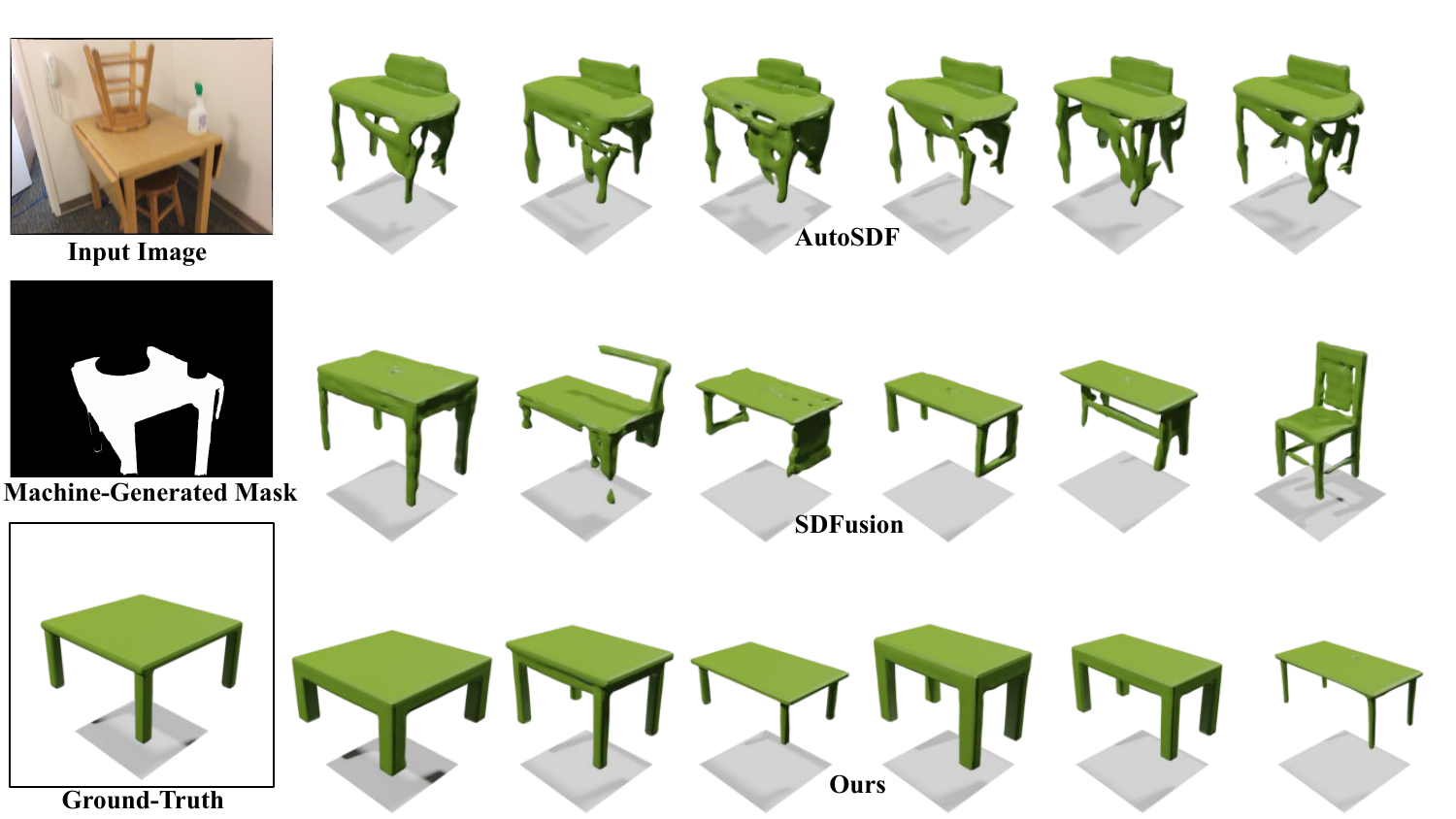}
    \caption{\textbf{More Qualitative Comparisons on Real-World Data.} Our method generates higher quality and more plausible hypotheses compared with other baselines.}
    \label{fig:viz_supp_real-world_3}
\end{figure}

\begin{figure}[htbp]
    \centering
    \includegraphics[width=1.0\linewidth]{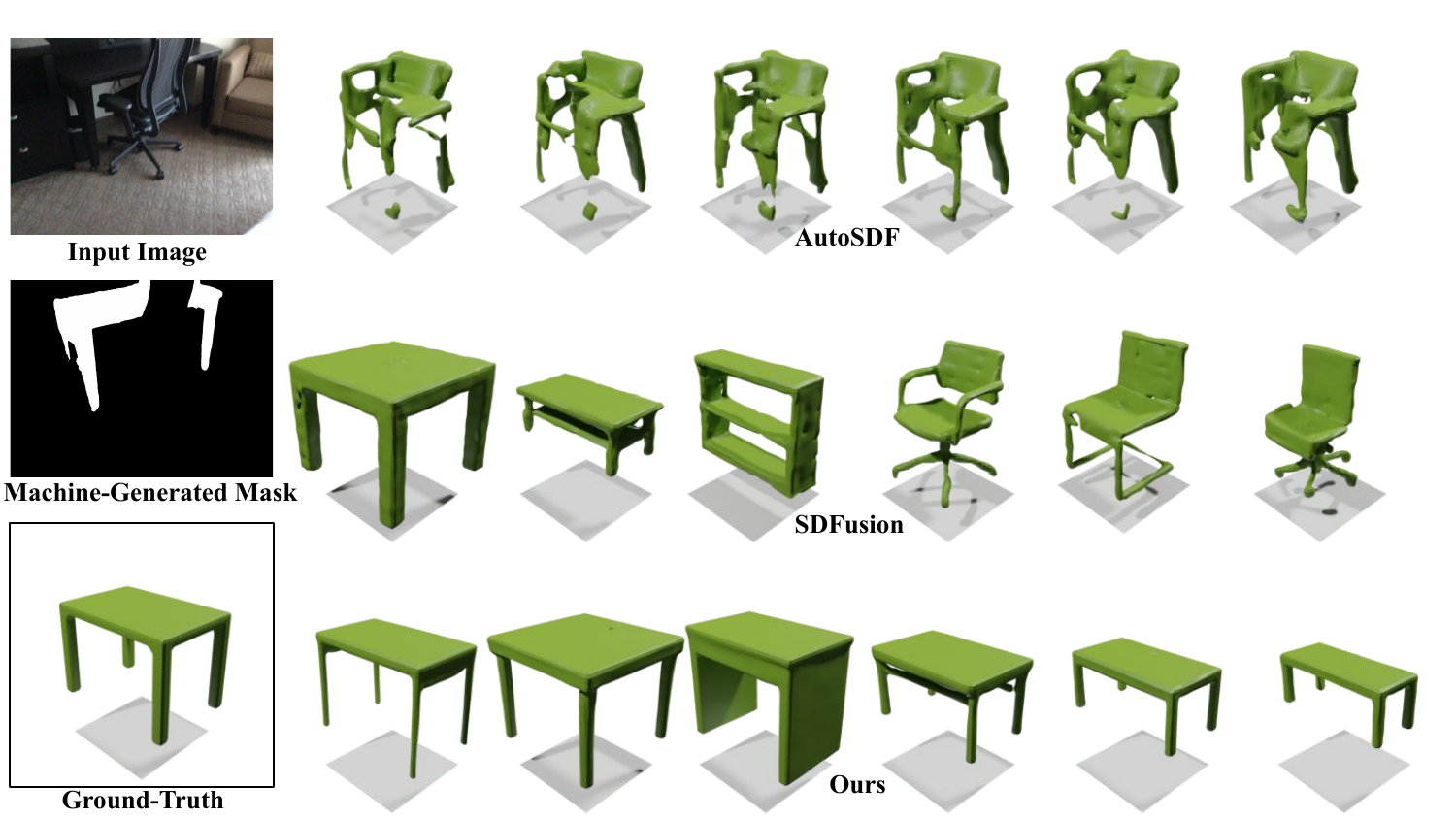}
    \hrule
    \includegraphics[width=1.0\linewidth]{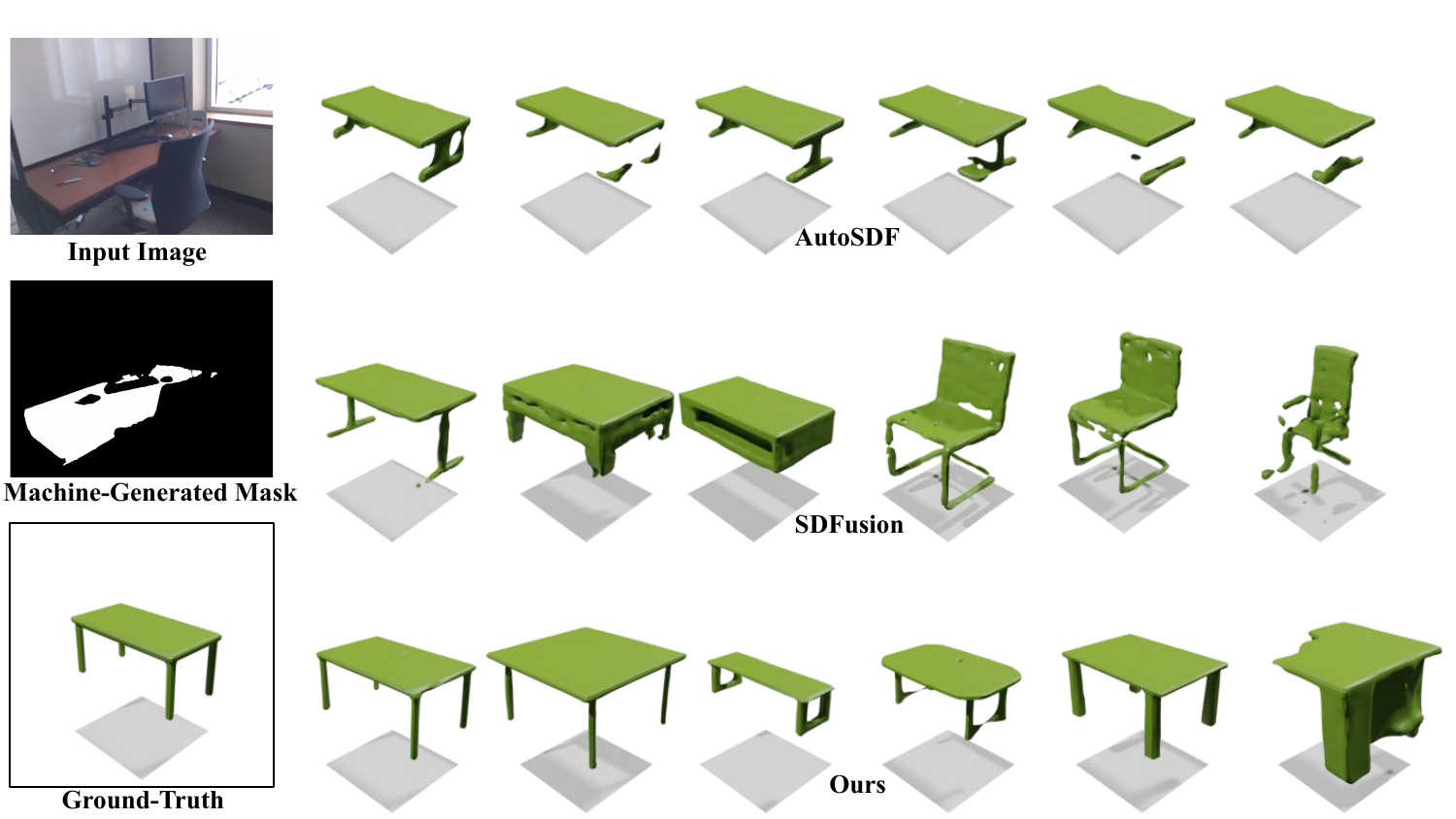}
    \caption{\textbf{More Qualitative Comparisons on Real-World Data.} Our method generates higher quality and more plausible hypotheses compared with other baselines.}
    \label{fig:viz_supp_real-world_3}
\end{figure}
\subsection{Further Implementation Details}
\label{subsec:further_details}
\subsubsection{Multi-Hypothesis Data Augmentation} 
In our approach, we allow the input image to be mapped to potentially multiple ground-truth shapes that align with the image. We initially classify CAD models from the dataset into similar groups. When evaluating two models, if they exhibit identical part counts and semantics, and their geometric similarity surpasses a predefined threshold, we classify them as similar. Then, for each rendered view of the target model, we extract per-pixel part labels and visible points in 3D space. We consider models from the same similar group as mapping candidates. We iterate through these candidates, employing exactly the same rendering parameters as the target model, and obtain their per-pixel part labels and visible parts in 3D space. We then compare this information with that of the target model: if the overlap of per-pixel part labels and the geometric similarity of visible parts exceed predefined thresholds, we include the candidate in the ground-truth mapping for the image.
\begin{figure}[htbp]
    \centering
    \includegraphics[width=1.0\linewidth]{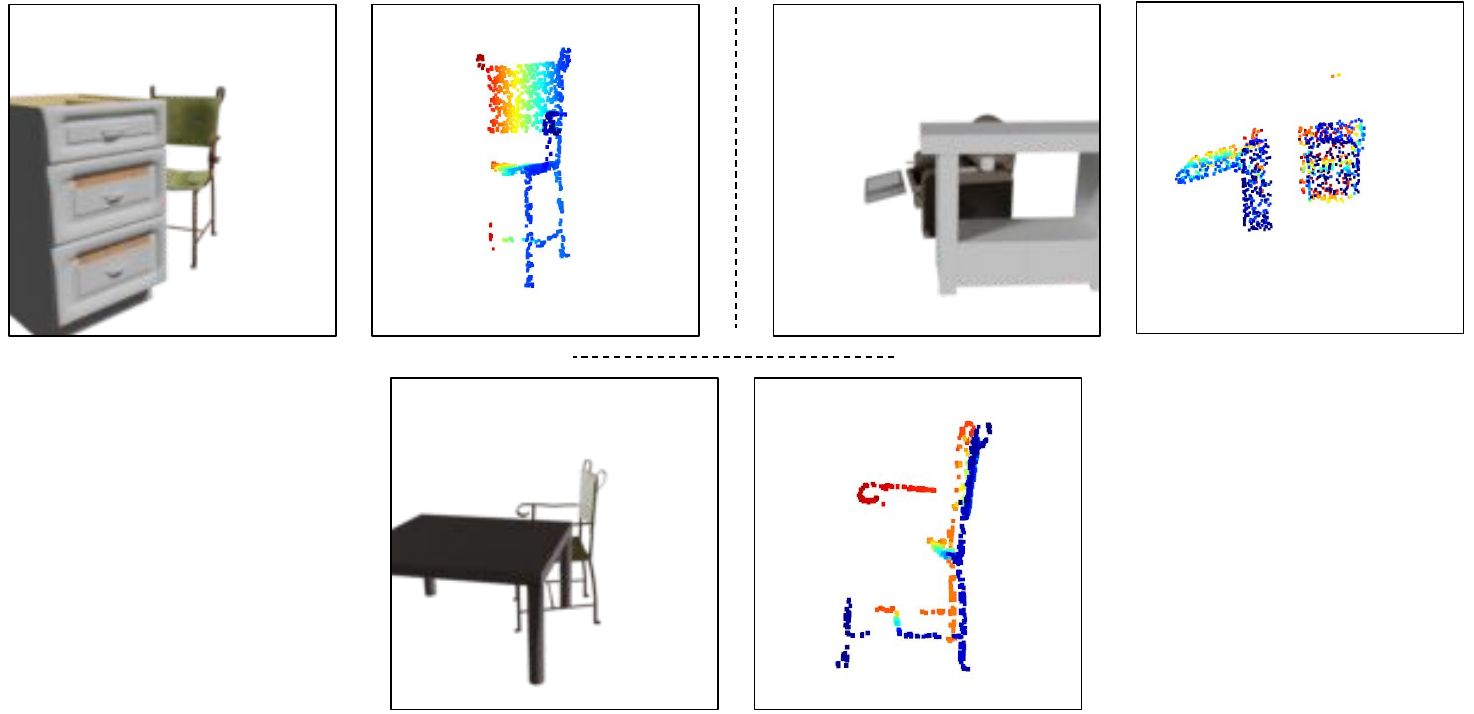}
    \caption{\textbf{Visible Points in 3D Space.} We present three sets of synthetic samples. Within each set, left is the rendering and right is the visible points of the target object in 3D space.}
    \label{fig:mapping_supp}
\end{figure}

\subsubsection{Network Architecture}

\noindent \textbf{Image Encoder.} We choose the "ViT-B/32" version of CLIP as our image encoder, yielding image encodings with a shape of $N'\times D'$, where $N'=50$ represents the number of image tokens, and $D'=768$ indicates the feature dimension.

\noindent \textbf{Conditional Cross-Attention.} The input sequence is multiplied with a weight matrix $Q$. This matrix multiplication results in a sequence of \textit{queries} with the shape of $N''\times d$, where $N''$ is the length of the input sequence and $d$ is a predefined hidden dimension. Likewise, the image encodings are multiplied with weight matrices $V$ and $K$ independently, generating \textit{values} and \textit{keys} respectively. Both are in the shape of $N'\times d$. Subsequently, each \textit{query} performs dot product with each \textit{key}, generating corresponding attention scores $\alpha$:
\begin{equation}
    \alpha_{mn} = \frac{softmax(q_m\cdot k_n)}{\sqrt{d}}.
\end{equation}
where $q_m$ is the $m$-th \textit{query} and $k_n$ is the $n$-th \text{key}. Then for each cell $i$ of the input sequence, its embedding is replaced by the weighted sum of \textit{values}:
\begin{equation}
    {emb}_i = \sum_{j=0}^{N'-1}\alpha_{ij}\cdot v_j.
\end{equation}
where $v_j$ is $j$-th \textit{value}. In practice, we apply 8 multi-head attention heads and a hidden dimension of 512.

\noindent \textbf{Transformer.} The transformer comprises 12 encoder layers, each with 12 multi-head attention heads and a hidden dimension of 768. Notably, it does not contain a decoder, indicating that all attention layers are self-attention. The training within the transformer is done in parallel. We feed the attention mask with upper-triangular matrix of $-\infty$, and zeros on the diagonal to make sure the information do not leak from the future elements. We use fourier features for the positional embedding for all locations $i$ following Tancik \etal~\cite{tancik2020fourier}.

\subsubsection{Fine-Tuning on Real-World Data}
To address the domain gap between our synthetic training pairs and real-world images, we fine-tune our pretrained model using real-world images from ScanNet. To preserve the ability for generating diverse shapes, we freeze the transformer-based generation backbone and only fine-tune the CLIP encoder and the conditional cross-attention module. We fix the batch size to 10 and utilize an initial learning rate of 5e-6 for the CLIP encoder and 1e-5 for the conditional cross-attention module. We fine-tune them for around 1,000 iterations.
\end{document}